\documentclass[letterpaper]{article}
\usepackage[preprint]{aaai2027}
\usepackage[hyphens]{url}
\usepackage{graphicx}
\usepackage{natbib}
\usepackage{caption}
\usepackage{booktabs}
\usepackage{amsmath}
\usepackage{array}
\usepackage{multirow}
\usepackage[table]{xcolor}
\usepackage{placeins}
\definecolor{rowhl}{HTML}{E8F0FB}   
\definecolor{grouphl}{HTML}{F4F6F8} 
\newcommand{\best}[1]{\textbf{#1}}          
\newcommand{\secondbest}[1]{\underline{#1}} 
\usepackage{pgfplots}
\pgfplotsset{compat=1.17}

\newcommand{\CEightStatus}{FAIL\_GATE}
\newcommand{\CEightWeights}{0.30/0.60/0.10}
\newcommand{\CEightTrain}{26/32 (81.2\%)}
\newcommand{\CEightTrainCalibrated}{24/32 (75.0\%)}
\newcommand{\CEightLOSO}{23/32 (71.9\%)}
\newcommand{\CEightDev}{11/15 (73.3\%)}
\newcommand{\CEightDevCalibrated}{11/15 (73.3\%)}
\newcommand{\CEightExtension}{84/114 (73.7\%)}
\newcommand{\CEightExtensionCalibrated}{84/114 (73.7\%)}
\newcommand{\CEightExtensionOriginal}{64/114 (56.1\%)}
\newcommand{\CEightGain}{0.0}
\newcommand{\CEightNonworse}{12/16}
\newcommand{\CEightBootstrapCI}{[-5.1, 5.0]}
\newcommand{\CEightConstraintRecords}{35}
\newcommand{\CEightPaidCalls}{0}

\newcommand{\metriccell}[2]{\shortstack{#1\\[-1pt]{\tiny #2}}}
\newcommand{\auditcell}[2]{\shortstack{#1\\[-1pt]{\tiny Wilson #2}}}
\newcommand{\ExtensionAllPairs}{135}

\newcommand{\IndependentAuditPairs}{32}
\newcommand{\IndependentAuditSeries}{14}

\newcommand{\IndependentAuditCoverageUnits}{160}

\newcommand{\COneConsensus}{64.0 [52.9,74.2]}

\newcommand{\COneConstraint}{78.6 [62.4,91.0]}

\newcommand{\COneLocation}{62.9 [42.9,80.6]}

\newcommand{\COneFPR}{5.7 [0.0,18.2]}

\newcommand{\COnePlural}{15.0 [5.5,26.7]}

\newcommand{\COnePluralWilson}{[8.1, 26.1]}
\newcommand{\COneReason}{41.7 [16.7,69.2]}

\newcommand{\COneReasonWilson}{[19.3, 68.0]}
\newcommand{\COneEvidence}{33.3 [8.3,63.6]}

\newcommand{\COneEvidenceWilson}{[13.8, 60.9]}
\newcommand{\CTwoConsensus}{53.5 [47.1,61.5]}

\newcommand{\CTwoConstraint}{74.0 [57.9,86.8]}

\newcommand{\CTwoLocation}{57.1 [36.4,77.1]}

\newcommand{\CTwoFPR}{11.4 [0.0,28.6]}

\newcommand{\CTwoPlural}{25.0 [14.0,38.0]}

\newcommand{\CTwoPluralWilson}{[15.8, 37.2]}
\newcommand{\CTwoReason}{66.7 [41.7,90.9]}

\newcommand{\CTwoReasonWilson}{[39.1, 86.2]}
\newcommand{\CTwoEvidence}{50.0 [20.0,81.8]}

\newcommand{\CTwoEvidenceWilson}{[25.4, 74.6]}
\newcommand{\CThreeConsensus}{56.1 [45.5,66.1]}

\newcommand{\CThreeConstraint}{77.1 [61.7,88.0]}

\newcommand{\CThreeLocation}{57.1 [41.7,71.9]}

\newcommand{\CThreeFPR}{0.0 [0.0,0.0]}

\newcommand{\CThreePlural}{28.3 [16.0,40.0]}

\newcommand{\CThreePluralWilson}{[18.5, 40.8]}
\newcommand{\CThreeReason}{83.3 [61.5,100.0]}

\newcommand{\CThreeReasonWilson}{[55.2, 95.3]}
\newcommand{\CThreeEvidence}{75.0 [41.7,100.0]}

\newcommand{\CThreeEvidenceWilson}{[46.8, 91.1]}
\newcommand{\CFourConsensus}{56.1 [48.7,64.2]}

\newcommand{\CFourConstraint}{71.7 [57.7,84.8]}

\newcommand{\CFourLocation}{57.1 [38.9,75.8]}

\newcommand{\CFourFPR}{20.0 [0.0,42.1]}

\newcommand{\CFourPlural}{31.7 [20.0,43.3]}

\newcommand{\CFourPluralWilson}{[21.3, 44.2]}
\newcommand{\CFourReason}{100.0 [100.0,100.0]}

\newcommand{\CFourReasonWilson}{[75.8, 100.0]}
\newcommand{\CFourEvidence}{83.3 [55.6,100.0]}

\newcommand{\CFourEvidenceWilson}{[55.2, 95.3]}

\newcommand{\CSixConsensus}{55.3 [45.6,65.2]}

\newcommand{\CSixConstraint}{70.1 [53.2,82.5]}

\newcommand{\CSixLocation}{51.4 [35.3,67.6]}

\newcommand{\CSixFPR}{0.0 [0.0,0.0]}

\newcommand{\CSixPlural}{23.3 [10.9,35.4]}

\newcommand{\CSixPluralWilson}{[14.4, 35.4]}
\newcommand{\CSixReason}{91.7 [70.0,100.0]}

\newcommand{\CSixReasonWilson}{[64.6, 98.5]}
\newcommand{\CSixEvidence}{66.7 [41.7,90.9]}

\newcommand{\CSixEvidenceWilson}{[39.1, 86.2]}

\newcommand{\CSevenCConsensus}{54.4 [44.5,64.0]}

\newcommand{\CSevenCConstraint}{56.5 [41.6,68.7]}

\newcommand{\CSevenCLocation}{40.0 [26.5,52.9]}

\newcommand{\CSevenCFPR}{5.7 [0.0,17.6]}

\newcommand{\CSevenCPlural}{36.7 [21.5,56.4]}

\newcommand{\CSevenCPluralWilson}{[25.6, 49.3]}
\newcommand{\CSevenCReason}{66.7 [38.5,92.9]}

\newcommand{\CSevenCReasonWilson}{[39.1, 86.2]}
\newcommand{\CSevenCEvidence}{41.7 [10.0,71.4]}

\newcommand{\CSevenCEvidenceWilson}{[19.3, 68.0]}

\newcommand{\IndCOneReason}{53.1 [34.3,73.3]}

\newcommand{\IndCOneReasonCount}{17/32}
\newcommand{\IndCOneReasonWilson}{[36.4, 69.1]}
\newcommand{\IndCOneEvidence}{50.0 [34.4,66.7]}

\newcommand{\IndCOneEvidenceCount}{16/32}
\newcommand{\IndCOneEvidenceWilson}{[33.6, 66.4]}
\newcommand{\IndCOneCoverage}{15.6 [9.4,22.6]}

\newcommand{\IndCOneCoverageCount}{25/160}
\newcommand{\IndCOneCoverageWilson}{[10.8, 22.0]}
\newcommand{\IndCOneAllReasonRate}{58.1}
\newcommand{\IndCOneAllEvidenceRate}{58.1}
\newcommand{\IndCOneAllCoverageRate}{23.9}
\newcommand{\IndCOneProfReasonRate}{59.4}
\newcommand{\IndCOneProfEvidenceRate}{61.5}
\newcommand{\IndCOneProfCoverageRate}{15.0}
\newcommand{\IndCOneNonProfReasonRate}{56.2}
\newcommand{\IndCOneNonProfEvidenceRate}{53.1}
\newcommand{\IndCOneNonProfCoverageRate}{37.2}
\newcommand{\IndCThreeReason}{78.1 [59.4,93.3]}

\newcommand{\IndCThreeReasonCount}{25/32}
\newcommand{\IndCThreeReasonWilson}{[61.2, 89.0]}
\newcommand{\IndCThreeEvidence}{78.1 [60.0,93.8]}

\newcommand{\IndCThreeEvidenceCount}{25/32}
\newcommand{\IndCThreeEvidenceWilson}{[61.2, 89.0]}
\newcommand{\IndCThreeCoverage}{36.9 [25.2,47.6]}

\newcommand{\IndCThreeCoverageCount}{59/160}
\newcommand{\IndCThreeCoverageWilson}{[29.8, 44.6]}
\newcommand{\IndCThreeAllReasonRate}{69.4}
\newcommand{\IndCThreeAllEvidenceRate}{69.4}
\newcommand{\IndCThreeAllCoverageRate}{44.4}
\newcommand{\IndCThreeProfReasonRate}{72.9}
\newcommand{\IndCThreeProfEvidenceRate}{74.0}
\newcommand{\IndCThreeProfCoverageRate}{37.5}
\newcommand{\IndCThreeNonProfReasonRate}{64.1}
\newcommand{\IndCThreeNonProfEvidenceRate}{62.5}
\newcommand{\IndCThreeNonProfCoverageRate}{54.7}
\newcommand{\IndCFourReason}{93.8 [82.8,100.0]}

\newcommand{\IndCFourReasonCount}{30/32}
\newcommand{\IndCFourReasonWilson}{[79.9, 98.3]}
\newcommand{\IndCFourEvidence}{90.6 [78.6,100.0]}

\newcommand{\IndCFourEvidenceCount}{29/32}
\newcommand{\IndCFourEvidenceWilson}{[75.8, 96.8]}
\newcommand{\IndCFourCoverage}{65.0 [58.8,70.0]}

\newcommand{\IndCFourCoverageCount}{104/160}
\newcommand{\IndCFourCoverageWilson}{[57.3, 72.0]}
\newcommand{\IndCFourAllReasonRate}{86.2}
\newcommand{\IndCFourAllEvidenceRate}{76.9}
\newcommand{\IndCFourAllCoverageRate}{62.1}
\newcommand{\IndCFourProfReasonRate}{92.7}
\newcommand{\IndCFourProfEvidenceRate}{81.2}
\newcommand{\IndCFourProfCoverageRate}{57.9}
\newcommand{\IndCFourNonProfReasonRate}{76.6}
\newcommand{\IndCFourNonProfEvidenceRate}{70.3}
\newcommand{\IndCFourNonProfCoverageRate}{68.4}

\newif\ifKimiCrossJudgeResults
\KimiCrossJudgeResultstrue
\newcommand{\CrossGLMCOneConsensus}{40.0}
\newcommand{\CrossGLMCOneConsensusCI}{[25.0, 57.7]}
\newcommand{\CrossGLMCOneConstraint}{78.6}
\newcommand{\CrossGLMCOneConstraintCI}{[62.4, 91.0]}
\newcommand{\CrossGLMCOneLocation}{62.9}
\newcommand{\CrossGLMCOneLocationCI}{[42.9, 80.6]}
\newcommand{\CrossGLMCOneFPR}{5.7}
\newcommand{\CrossGLMCOneFPRCI}{[0.0, 18.2]}
\newcommand{\CrossGLMCTwoConsensus}{64.0}
\newcommand{\CrossGLMCTwoConsensusCI}{[48.0, 80.0]}
\newcommand{\CrossGLMCTwoConstraint}{74.0}
\newcommand{\CrossGLMCTwoConstraintCI}{[57.9, 86.8]}
\newcommand{\CrossGLMCTwoLocation}{57.1}
\newcommand{\CrossGLMCTwoLocationCI}{[36.4, 77.1]}
\newcommand{\CrossGLMCTwoFPR}{11.4}
\newcommand{\CrossGLMCTwoFPRCI}{[0.0, 28.6]}
\newcommand{\CrossGLMCThreeConsensus}{56.0}
\newcommand{\CrossGLMCThreeConsensusCI}{[41.4, 72.7]}
\newcommand{\CrossGLMCThreeConstraint}{77.1}
\newcommand{\CrossGLMCThreeConstraintCI}{[61.7, 88.0]}
\newcommand{\CrossGLMCThreeLocation}{57.1}
\newcommand{\CrossGLMCThreeLocationCI}{[41.7, 71.9]}
\newcommand{\CrossGLMCThreeFPR}{0.0}
\newcommand{\CrossGLMCThreeFPRCI}{[0.0, 0.0]}
\newcommand{\CrossGLMCFourConsensus}{60.0}
\newcommand{\CrossGLMCFourConsensusCI}{[44.0, 77.3]}
\newcommand{\CrossGLMCFourConstraint}{71.7}
\newcommand{\CrossGLMCFourConstraintCI}{[57.7, 84.8]}
\newcommand{\CrossGLMCFourLocation}{57.1}
\newcommand{\CrossGLMCFourLocationCI}{[38.9, 75.8]}
\newcommand{\CrossGLMCFourFPR}{20.0}
\newcommand{\CrossGLMCFourFPRCI}{[0.0, 42.1]}
\newcommand{\CrossGLMCSixConsensus}{64.0}
\newcommand{\CrossGLMCSixConsensusCI}{[40.0, 81.2]}
\newcommand{\CrossGLMCSixConstraint}{70.1}
\newcommand{\CrossGLMCSixConstraintCI}{[53.2, 82.5]}
\newcommand{\CrossGLMCSixLocation}{51.4}
\newcommand{\CrossGLMCSixLocationCI}{[35.3, 67.6]}
\newcommand{\CrossGLMCSixFPR}{0.0}
\newcommand{\CrossGLMCSixFPRCI}{[0.0, 0.0]}
\newcommand{\CrossGLMCSevenCConsensus}{48.0}
\newcommand{\CrossGLMCSevenCConsensusCI}{[31.8, 63.0]}
\newcommand{\CrossGLMCSevenCConstraint}{56.5}
\newcommand{\CrossGLMCSevenCConstraintCI}{[41.6, 68.7]}
\newcommand{\CrossGLMCSevenCLocation}{40.0}
\newcommand{\CrossGLMCSevenCLocationCI}{[26.5, 52.9]}
\newcommand{\CrossGLMCSevenCFPR}{5.7}
\newcommand{\CrossGLMCSevenCFPRCI}{[0.0, 17.6]}
\newcommand{\CrossQwenCOneConsensus}{48.0}
\newcommand{\CrossQwenCOneConsensusCI}{[23.5, 72.0]}
\newcommand{\CrossQwenCOneConstraint}{79.9}
\newcommand{\CrossQwenCOneConstraintCI}{[68.9, 91.3]}
\newcommand{\CrossQwenCOneLocation}{60.0}
\newcommand{\CrossQwenCOneLocationCI}{[38.9, 81.2]}
\newcommand{\CrossQwenCOneFPR}{28.6}
\newcommand{\CrossQwenCOneFPRCI}{[6.1, 52.6]}
\newcommand{\CrossQwenCTwoConsensus}{64.0}
\newcommand{\CrossQwenCTwoConsensusCI}{[40.9, 84.0]}
\newcommand{\CrossQwenCTwoConstraint}{69.4}
\newcommand{\CrossQwenCTwoConstraintCI}{[64.1, 75.6]}
\newcommand{\CrossQwenCTwoLocation}{17.1}
\newcommand{\CrossQwenCTwoLocationCI}{[0.0, 36.4]}
\newcommand{\CrossQwenCTwoFPR}{82.9}
\newcommand{\CrossQwenCTwoFPRCI}{[63.6, 100.0]}
\newcommand{\CrossQwenCThreeConsensus}{40.0}
\newcommand{\CrossQwenCThreeConsensusCI}{[22.9, 60.0]}
\newcommand{\CrossQwenCThreeConstraint}{73.9}
\newcommand{\CrossQwenCThreeConstraintCI}{[66.8, 82.1]}
\newcommand{\CrossQwenCThreeLocation}{37.1}
\newcommand{\CrossQwenCThreeLocationCI}{[16.2, 59.5]}
\newcommand{\CrossQwenCThreeFPR}{51.4}
\newcommand{\CrossQwenCThreeFPRCI}{[25.7, 75.8]}
\newcommand{\CrossQwenCFourConsensus}{36.0}
\newcommand{\CrossQwenCFourConsensusCI}{[22.7, 52.2]}
\newcommand{\CrossQwenCFourConstraint}{77.3}
\newcommand{\CrossQwenCFourConstraintCI}{[70.9, 85.7]}
\newcommand{\CrossQwenCFourLocation}{40.0}
\newcommand{\CrossQwenCFourLocationCI}{[18.9, 63.6]}
\newcommand{\CrossQwenCFourFPR}{54.3}
\newcommand{\CrossQwenCFourFPRCI}{[27.3, 77.8]}
\newcommand{\CrossQwenCSixConsensus}{36.0}
\newcommand{\CrossQwenCSixConsensusCI}{[19.2, 55.6]}
\newcommand{\CrossQwenCSixConstraint}{82.1}
\newcommand{\CrossQwenCSixConstraintCI}{[72.3, 92.0]}
\newcommand{\CrossQwenCSixLocation}{57.1}
\newcommand{\CrossQwenCSixLocationCI}{[35.3, 78.1]}
\newcommand{\CrossQwenCSixFPR}{31.4}
\newcommand{\CrossQwenCSixFPRCI}{[10.8, 55.6]}
\newcommand{\CrossQwenCSevenCConsensus}{60.0}
\newcommand{\CrossQwenCSevenCConsensusCI}{[42.9, 75.0]}
\newcommand{\CrossQwenCSevenCConstraint}{67.5}
\newcommand{\CrossQwenCSevenCConstraintCI}{[57.1, 77.8]}
\newcommand{\CrossQwenCSevenCLocation}{34.3}
\newcommand{\CrossQwenCSevenCLocationCI}{[17.6, 51.5]}
\newcommand{\CrossQwenCSevenCFPR}{34.3}
\newcommand{\CrossQwenCSevenCFPRCI}{[11.8, 58.3]}
\newcommand{\CrossKimiCOneConsensus}{48.0}
\newcommand{\CrossKimiCOneConsensusCI}{[25.9, 71.4]}
\newcommand{\CrossKimiCOneConstraint}{83.0}
\newcommand{\CrossKimiCOneConstraintCI}{[72.4, 92.0]}
\newcommand{\CrossKimiCOneLocation}{54.3}
\newcommand{\CrossKimiCOneLocationCI}{[40.0, 68.6]}
\newcommand{\CrossKimiCOneFPR}{8.6}
\newcommand{\CrossKimiCOneFPRCI}{[0.0, 25.0]}
\newcommand{\CrossKimiCTwoConsensus}{56.0}
\newcommand{\CrossKimiCTwoConsensusCI}{[36.0, 76.2]}
\newcommand{\CrossKimiCTwoConstraint}{82.6}
\newcommand{\CrossKimiCTwoConstraintCI}{[72.4, 92.4]}
\newcommand{\CrossKimiCTwoLocation}{57.1}
\newcommand{\CrossKimiCTwoLocationCI}{[37.1, 76.5]}
\newcommand{\CrossKimiCTwoFPR}{25.7}
\newcommand{\CrossKimiCTwoFPRCI}{[5.7, 50.0]}
\newcommand{\CrossKimiCThreeConsensus}{68.0}
\newcommand{\CrossKimiCThreeConsensusCI}{[52.8, 85.7]}
\newcommand{\CrossKimiCThreeConstraint}{80.1}
\newcommand{\CrossKimiCThreeConstraintCI}{[72.1, 88.0]}
\newcommand{\CrossKimiCThreeLocation}{51.4}
\newcommand{\CrossKimiCThreeLocationCI}{[35.1, 68.6]}
\newcommand{\CrossKimiCThreeFPR}{14.3}
\newcommand{\CrossKimiCThreeFPRCI}{[0.0, 34.3]}
\newcommand{\CrossKimiCFourConsensus}{52.0}
\newcommand{\CrossKimiCFourConsensusCI}{[33.3, 72.0]}
\newcommand{\CrossKimiCFourConstraint}{81.2}
\newcommand{\CrossKimiCFourConstraintCI}{[73.1, 88.6]}
\newcommand{\CrossKimiCFourLocation}{62.9}
\newcommand{\CrossKimiCFourLocationCI}{[47.1, 78.1]}
\newcommand{\CrossKimiCFourFPR}{11.4}
\newcommand{\CrossKimiCFourFPRCI}{[0.0, 29.4]}
\newcommand{\CrossKimiCSixConsensus}{52.0}
\newcommand{\CrossKimiCSixConsensusCI}{[30.8, 75.0]}
\newcommand{\CrossKimiCSixConstraint}{78.5}
\newcommand{\CrossKimiCSixConstraintCI}{[67.4, 89.2]}
\newcommand{\CrossKimiCSixLocation}{54.3}
\newcommand{\CrossKimiCSixLocationCI}{[37.8, 70.6]}
\newcommand{\CrossKimiCSixFPR}{14.3}
\newcommand{\CrossKimiCSixFPRCI}{[0.0, 35.1]}
\newcommand{\CrossKimiCSevenCConsensus}{72.0}
\newcommand{\CrossKimiCSevenCConsensusCI}{[52.2, 88.9]}
\newcommand{\CrossKimiCSevenCConstraint}{41.4}
\newcommand{\CrossKimiCSevenCConstraintCI}{[28.3, 53.8]}
\newcommand{\CrossKimiCSevenCLocation}{34.3}
\newcommand{\CrossKimiCSevenCLocationCI}{[22.9, 45.7]}
\newcommand{\CrossKimiCSevenCFPR}{11.4}
\newcommand{\CrossKimiCSevenCFPRCI}{[0.0, 29.4]}

\newcommand{\DeltaStructureConsensus}{+0.0}

\newcommand{\DeltaFlatHLConsensus}{-0.9}

\newcommand{\DeltaDetailedConsensus}{+0.9}

\newcommand{\DeltaCoarseConsensus}{-1.8}

\newcommand{\RichCOneSuccess}{64/64}
\newcommand{\RichCOneCompletionMedian}{1222.5}
\newcommand{\RichCOneCompletionIQR}{900.8--1636.5}

\newcommand{\RichCOneSpansMedian}{3}
\newcommand{\RichCOneSpansIQR}{3--3}
\newcommand{\RichCOneCharsMedian}{76}
\newcommand{\RichCOneCharsIQR}{61.8--87}

\newcommand{\RichCThreeSuccess}{64/64}
\newcommand{\RichCThreeCompletionMedian}{3165.5}
\newcommand{\RichCThreeCompletionIQR}{2589--4060}

\newcommand{\RichCThreeSpansMedian}{16}
\newcommand{\RichCThreeSpansIQR}{16--16}
\newcommand{\RichCThreeCharsMedian}{435}
\newcommand{\RichCThreeCharsIQR}{329.5--720.5}

\newcommand{\RichCFourSuccess}{64/64}
\newcommand{\RichCFourCompletionMedian}{3427.5}
\newcommand{\RichCFourCompletionIQR}{2816.2--4290.8}

\newcommand{\RichCFourSpansMedian}{16}
\newcommand{\RichCFourSpansIQR}{16--16}
\newcommand{\RichCFourCharsMedian}{778.5}
\newcommand{\RichCFourCharsIQR}{673--964.5}

\newcommand{\CFourConstraintFOne}{74.0}
\newcommand{\CFourDetectLocate}{60.0}
\newcommand{\CFourFPRVTwo}{13.3}
\newcommand{\CFourINineFOne}{64.5}
\newcommand{\CFourINineLocate}{50.0}
\newcommand{\CFourINineSensitivity}{51.9}
\newcommand{\CFourINineSpecificity}{90.7}
\newcommand{\CFourISevenFOne}{83.4}
\newcommand{\CFourISevenLocate}{70.6}
\newcommand{\CFourISevenSensitivity}{84.3}
\newcommand{\CFourISevenSpecificity}{82.4}
\newcommand{\CFourLightKappa}{0.45}
\newcommand{\CFourNegativeAgreement}{77.5}
\newcommand{\CFourPABAK}{0.47}
\newcommand{\CFourPositiveAgreement}{67.1}
\newcommand{\CFourPrefAgreement}{31.6}
\newcommand{\CFourPrefHigh}{56.1}
\newcommand{\CFourPrefKappa}{0.27}
\newcommand{\CFourPrefLow}{51.8}
\newcommand{\CFourPrefMajority}{50.9}
\newcommand{\CFourPrefMean}{53.8}

\newcommand{\CFourRunOneBalanced}{75.7}

\newcommand{\CFourRunOneFOne}{71.7}
\newcommand{\CFourRunOneFPR}{11.4}

\newcommand{\CFourRunOneLocate}{54.3}
\newcommand{\CFourRunOnePref}{51.8}
\newcommand{\CFourRunOneSensitivity}{62.9}
\newcommand{\CFourRunOneSpecificity}{88.6}
\newcommand{\CFourRunTwoBalanced}{81.4}

\newcommand{\CFourRunTwoFOne}{78.6}
\newcommand{\CFourRunTwoFPR}{8.6}

\newcommand{\CFourRunTwoLocate}{68.6}
\newcommand{\CFourRunTwoPref}{53.5}
\newcommand{\CFourRunTwoSensitivity}{71.4}
\newcommand{\CFourRunTwoSpecificity}{91.4}
\newcommand{\CFourRunZeroBalanced}{74.3}

\newcommand{\CFourRunZeroFOne}{71.7}
\newcommand{\CFourRunZeroFPR}{20.0}

\newcommand{\CFourRunZeroLocate}{57.1}
\newcommand{\CFourRunZeroPref}{56.1}
\newcommand{\CFourRunZeroSensitivity}{68.6}
\newcommand{\CFourRunZeroSpecificity}{80.0}
\newcommand{\CFourSensitivity}{67.6}
\newcommand{\CFourSpecificity}{86.7}
\newcommand{\CFourTauPointFiveAccuracy}{34.8}
\newcommand{\CFourTauPointFiveTie}{58.2}
\newcommand{\CFourTauPointOneAccuracy}{65.2}
\newcommand{\CFourTauPointOneTie}{15.5}
\newcommand{\CFourTauPointTwoFiveAccuracy}{53.8}
\newcommand{\CFourTauPointTwoFiveTie}{35.4}
\newcommand{\CFourTauZeroAccuracy}{71.9}
\newcommand{\CFourTauZeroTie}{2.3}

\newcommand{\COneCFourCIHigh}{13.4}
\newcommand{\COneCFourCILow}{-9.9}
\newcommand{\COneCFourDelta}{1.5}
\newcommand{\COneCFourHolm}{1.000}
\newcommand{\COneCThreeCIHigh}{10.3}
\newcommand{\COneCThreeCILow}{-11.2}
\newcommand{\COneCThreeDelta}{-0.5}
\newcommand{\COneCThreeHolm}{1.000}
\newcommand{\COneConstraintFOne}{76.9}
\newcommand{\COneDetectLocate}{61.0}
\newcommand{\COneFPRVTwo}{5.7}
\newcommand{\COneINineFOne}{66.6}
\newcommand{\COneINineLocate}{44.4}
\newcommand{\COneINineSensitivity}{53.7}
\newcommand{\COneINineSpecificity}{92.6}
\newcommand{\COneISevenFOne}{87.2}
\newcommand{\COneISevenLocate}{78.4}
\newcommand{\COneISevenSensitivity}{80.4}
\newcommand{\COneISevenSpecificity}{96.1}
\newcommand{\COneLightKappa}{0.63}
\newcommand{\COneNegativeAgreement}{86.6}
\newcommand{\COnePABAK}{0.66}
\newcommand{\COnePositiveAgreement}{76.3}
\newcommand{\COnePrefAgreement}{32.5}
\newcommand{\COnePrefHigh}{64.0}
\newcommand{\COnePrefKappa}{0.31}
\newcommand{\COnePrefLow}{50.0}
\newcommand{\COnePrefMajority}{58.8}
\newcommand{\COnePrefMean}{55.3}

\newcommand{\COneRunOneBalanced}{81.4}

\newcommand{\COneRunOneFOne}{77.3}
\newcommand{\COneRunOneFPR}{2.9}

\newcommand{\COneRunOneLocate}{62.9}
\newcommand{\COneRunOnePref}{50.0}
\newcommand{\COneRunOneSensitivity}{65.7}
\newcommand{\COneRunOneSpecificity}{97.1}
\newcommand{\COneRunTwoBalanced}{78.6}

\newcommand{\COneRunTwoFOne}{74.8}
\newcommand{\COneRunTwoFPR}{8.6}

\newcommand{\COneRunTwoLocate}{57.1}
\newcommand{\COneRunTwoPref}{51.8}
\newcommand{\COneRunTwoSensitivity}{65.7}
\newcommand{\COneRunTwoSpecificity}{91.4}
\newcommand{\COneRunZeroBalanced}{81.4}

\newcommand{\COneRunZeroFOne}{78.6}
\newcommand{\COneRunZeroFPR}{5.7}

\newcommand{\COneRunZeroLocate}{62.9}
\newcommand{\COneRunZeroPref}{64.0}
\newcommand{\COneRunZeroSensitivity}{68.6}
\newcommand{\COneRunZeroSpecificity}{94.3}
\newcommand{\COneSensitivity}{66.7}
\newcommand{\COneSpecificity}{94.3}
\newcommand{\COneTauPointFiveAccuracy}{47.7}
\newcommand{\COneTauPointFiveTie}{42.7}
\newcommand{\COneTauPointOneAccuracy}{55.3}
\newcommand{\COneTauPointOneTie}{31.0}
\newcommand{\COneTauPointTwoFiveAccuracy}{55.3}
\newcommand{\COneTauPointTwoFiveTie}{31.0}
\newcommand{\COneTauZeroAccuracy}{55.3}
\newcommand{\COneTauZeroTie}{30.7}

\newcommand{\CThreeCFourCIHigh}{13.3}
\newcommand{\CThreeCFourCILow}{-9.3}
\newcommand{\CThreeCFourDelta}{2.0}
\newcommand{\CThreeCFourHolm}{1.000}
\newcommand{\CThreeConstraintFOne}{74.5}
\newcommand{\CThreeDetectLocate}{55.2}
\newcommand{\CThreeFPRVTwo}{10.5}
\newcommand{\CThreeINineFOne}{68.5}
\newcommand{\CThreeINineLocate}{46.3}
\newcommand{\CThreeINineSensitivity}{55.6}
\newcommand{\CThreeINineSpecificity}{92.6}
\newcommand{\CThreeISevenFOne}{80.5}
\newcommand{\CThreeISevenLocate}{64.7}
\newcommand{\CThreeISevenSensitivity}{76.5}
\newcommand{\CThreeISevenSpecificity}{86.3}
\newcommand{\CThreeLightKappa}{0.52}
\newcommand{\CThreeNegativeAgreement}{81.4}
\newcommand{\CThreePABAK}{0.54}
\newcommand{\CThreePositiveAgreement}{70.1}
\newcommand{\CThreePrefAgreement}{29.8}
\newcommand{\CThreePrefHigh}{57.0}
\newcommand{\CThreePrefKappa}{0.25}
\newcommand{\CThreePrefLow}{55.3}
\newcommand{\CThreePrefMajority}{61.4}
\newcommand{\CThreePrefMean}{55.8}

\newcommand{\CThreeRunOneBalanced}{72.9}

\newcommand{\CThreeRunOneFOne}{70.6}
\newcommand{\CThreeRunOneFPR}{20.0}

\newcommand{\CThreeRunOneLocate}{48.6}
\newcommand{\CThreeRunOnePref}{57.0}
\newcommand{\CThreeRunOneSensitivity}{65.7}
\newcommand{\CThreeRunOneSpecificity}{80.0}
\newcommand{\CThreeRunTwoBalanced}{78.6}

\newcommand{\CThreeRunTwoFOne}{75.7}
\newcommand{\CThreeRunTwoFPR}{11.4}

\newcommand{\CThreeRunTwoLocate}{60.0}
\newcommand{\CThreeRunTwoPref}{55.3}
\newcommand{\CThreeRunTwoSensitivity}{68.6}
\newcommand{\CThreeRunTwoSpecificity}{88.6}
\newcommand{\CThreeRunZeroBalanced}{81.4}

\newcommand{\CThreeRunZeroFOne}{77.1}
\newcommand{\CThreeRunZeroFPR}{0.0}

\newcommand{\CThreeRunZeroLocate}{57.1}
\newcommand{\CThreeRunZeroPref}{55.3}
\newcommand{\CThreeRunZeroSensitivity}{62.9}
\newcommand{\CThreeRunZeroSpecificity}{100.0}
\newcommand{\CThreeSensitivity}{65.7}
\newcommand{\CThreeSpecificity}{89.5}
\newcommand{\CThreeTauPointFiveAccuracy}{37.1}
\newcommand{\CThreeTauPointFiveTie}{54.7}
\newcommand{\CThreeTauPointOneAccuracy}{66.1}
\newcommand{\CThreeTauPointOneTie}{14.9}
\newcommand{\CThreeTauPointTwoFiveAccuracy}{55.8}
\newcommand{\CThreeTauPointTwoFiveTie}{30.1}
\newcommand{\CThreeTauZeroAccuracy}{71.6}
\newcommand{\CThreeTauZeroTie}{5.0}
\newcommand{\ConstraintPairsVTwo}{35}

\newcommand{\DirectGLMAgreement}{70.2}
\newcommand{\DirectGLMConstraintFOne}{89.5}
\newcommand{\DirectGLMDetectLocate}{81.9}
\newcommand{\DirectGLMFPR}{1.9}
\newcommand{\DirectGLMKappa}{0.60}
\newcommand{\DirectGLMPreference}{85.1}
\newcommand{\DirectGLMRunOneFOne}{91.9}

\newcommand{\DirectGLMRunOnePreference}{85.1}
\newcommand{\DirectGLMRunTwoFOne}{90.4}

\newcommand{\DirectGLMRunTwoPreference}{85.1}
\newcommand{\DirectGLMRunZeroFOne}{86.2}

\newcommand{\DirectGLMRunZeroPreference}{85.1}

\newcommand{\DirectQwenConstraintFOne}{83.0}
\newcommand{\DirectQwenDetectLocate}{51.4}
\newcommand{\DirectQwenFPR}{17.1}
\newcommand{\DirectQwenPreference}{80.7}

\newcommand{\FiveFivePairs}{81}
\newcommand{\FourFivePairs}{33}
\newcommand{\GEvalGLMConstraintFOne}{43.2}
\newcommand{\GEvalGLMDetectLocate}{28.6}
\newcommand{\GEvalGLMFPR}{2.9}
\newcommand{\GEvalGLMPreference}{78.9}

\newcommand{\GEvalGLMSensitivity}{28.6}
\newcommand{\GEvalGLMSpecificity}{97.1}
\newcommand{\GEvalTauPointFiveAccuracy}{78.9}
\newcommand{\GEvalTauPointFiveTie}{14.9}
\newcommand{\GEvalTauPointOneAccuracy}{78.9}
\newcommand{\GEvalTauPointOneTie}{14.9}
\newcommand{\GEvalTauPointTwoFiveAccuracy}{78.9}
\newcommand{\GEvalTauPointTwoFiveTie}{14.9}
\newcommand{\GEvalTauZeroAccuracy}{78.9}
\newcommand{\GEvalTauZeroTie}{14.9}

\newcommand{\PanelSupermajorityPairs}{114}
\newcommand{\RepeatSeries}{16}

\newcommand{\ThreeFiveLeaveRetention}{40.0}
\newcommand{\ThreeFiveLeaveUnresolved}{60.0}
\newcommand{\ThreeFivePairs}{21}
\newcommand{\ThreeFiveTripleRetention}{70.0}
\newcommand{\ThreeFiveTripleReversal}{30.0}
\newcommand{\WithinPanelDisagreementPairs}{21}

\newcommand{\method}{\textsc{DramaGrow}}
\newcommand{\verifiable}{\textsc{Verifiable}}
\newcommand{\panelsuper}{\textsc{Panel-Supermajority}}
\newcommand{\withindisagreement}{\textsc{Within-Panel Disagreement}}

\title{DramaGrow: A Regime-Aware Benchmark for Auditing LLM Judges\\of Multi-Episode Drama}
\author{Hui Zhang\textsuperscript{1}, Yibo Su\textsuperscript{2}}
\affiliations{\textsuperscript{1}Beijing Institute of Technology\\\textsuperscript{2}Heilongjiang University\\\texttt{huizhang.ai.2025@gmail.com}}

\begin{document}
\maketitle

\begin{abstract}
LLMs can now draft complete multi-episode dramas, and LLM judges are increasingly used to evaluate them. Assessing the reliability of these judges is difficult because drama quality encompasses specification-backed errors, reproducible preferences, and legitimate disagreement among human raters. Most existing evaluations reduce these cases to a single score from one judge call and leave tie calibration, call-to-call variation, and the vote margin underlying the human reference unreported. We introduce \method{}, a regime-aware benchmark and audit protocol for AI-generated Chinese five-episode drama. The benchmark assigns verifiable constraint violations, four-of-five preferences, and within-panel disagreements to separate denominators and evidence rules. It contains 800 episodes, 120 byte-preserving interventions, and 1{,}155 blinded five-rater records, including a frozen extension with \PanelSupermajorityPairs{} supermajority pairs and \WithinPanelDisagreementPairs{} disagreement pairs. On the frozen sets of \PanelSupermajorityPairs{} preference pairs and \ConstraintPairsVTwo{} constraint interventions, the repeated absolute-score rubrics yield overlapping performance, with mean accuracies of \CFourPrefMean--\CThreePrefMean\% and three-call agreement of \CThreePrefAgreement--\COnePrefAgreement\%, and their accuracy falls from \CThreeTauZeroAccuracy--\CFourTauZeroAccuracy\% to \CFourTauPointFiveAccuracy--\CThreeTauPointFiveAccuracy\% as the tie band widens. DirectPair--GLM reaches \DirectGLMPreference\% preference accuracy, \DirectGLMConstraintFOne\% constraint F1, and \DirectGLMDetectLocate\% Detect+Locate on the same items, and exact subpanel enumeration identifies four-of-five as the smallest stable human vote margin. Evaluation protocol, tie calibration, judge-call variation, and panel margin therefore affect performance as much as rubric design. We will release the specification packages, scripts, interventions, anonymized human votes and rationales, frozen judge runs, and analysis code.
\end{abstract}

\section{Introduction}

AI systems can generate scenes, episodes, and increasingly complete serial narratives, and studios that produce vertical short-form dramas must decide which machine-generated drafts to retain, revise, or discard. Evaluating a five-episode drama requires reasoning at several structural scales: dialogue may sound unnatural within a scene, an episode may fail to escalate or resolve a setup, and a later episode may contradict an earlier fact or allow a character to act on information never disclosed to them. Human editors devote substantial effort to identifying these failures, which has motivated the use of LLM judges. This paper examines the reliability of such judges and introduces a benchmark for reproducible and auditable evaluation.

Current evaluations often compare a holistic score from a single judge call against an aggregated human label. Prior work has tested story metrics through perturbations and human judgments \citep{guan-etal-2021-openmeva}, evaluated drama continuation along domain-specific dimensions \citep{ma2025dramabench}, learned predictors of creative-writing preferences \citep{fein-etal-2026-litbench}, and built LLM evaluators around decomposed criteria \citep{liu-etal-2023-g,hashemi-etal-2024-llm}. These pipelines typically leave evidence locations unspecified, report only one judge call, and aggregate human votes without retaining the vote margin.

Four factors complicate the evaluation of drama judges. (i) Mixed evaluation regimes: some defects can be verified against a story specification, some qualities produce reproducible panel preferences, and some comparisons remain genuinely disputed. A single labeling rule cannot represent all three cases. (ii) Judge-call variation: repeated calls on the same input can produce different decisions. (iii) Tie calibration: absolute-score protocols require a tie threshold, and both accuracy and system ranking vary with this threshold. Clean controls are also necessary to identify judges that overpredict defects. (iv) Human-reference construction: panel size, vote margin, and panel composition determine the stability of reference labels. Rubric adaptation can also expose test information when the data split is not sealed.

\method{} measures each factor directly. A fixed five-rater panel labels blinded version pairs and records votes, reasons, and evidence spans. These annotations define the reference labels, validate the controlled corruptions, and support a subsequent audit of judge rationales. The data are divided into three regimes: \verifiable{} defects checked against the specification, \panelsuper{} preferences supported by at least four of five raters, and \withindisagreement{} pairs for which the full vote distributions are retained. Judges return scores, fatal-gate decisions, and localized evidence. We evaluate the three core rubric configurations in three independent runs on identical inputs. We also pair each corrupted script with its clean source, preregister and sweep the tie thresholds, evaluate every subpanel, and seal the rubrics, tasks, mappings, and analysis code before test execution.

We evaluate scoring protocol, tie calibration, rubric structure, judge-call variation, and panel margin on common evaluation sets. Our contributions are as follows:
\begin{itemize}\setlength{\itemsep}{1pt}\setlength{\parskip}{0pt}\setlength{\topsep}{2pt}
\item A regime-aware benchmark for auditing drama judges, which assigns verifiable constraints, panel-supermajority preferences, and within-panel disagreements to separate analysis rules over 800 generated episodes, 120 byte-preserving interventions, and 1{,}155 blinded five-rater records with reasons and evidence.
\item An audit protocol combining repeated identical calls, preregistered tie-threshold sweeps, class-balanced fatal tests, localization credit, exact subpanel enumeration, and two-level bootstrap inference.
\item The finding that evaluation protocol has a larger measured effect than rubric representation: DirectPair--GLM reaches \DirectGLMPreference\% preference accuracy and \DirectGLMConstraintFOne\% constraint F1, the three absolute-score rubrics yield overlapping repeated-call results, and four-of-five is the smallest stable panel margin.
\end{itemize}

\begin{figure*}[t]
  \centering
  \includegraphics[width=0.92\textwidth]{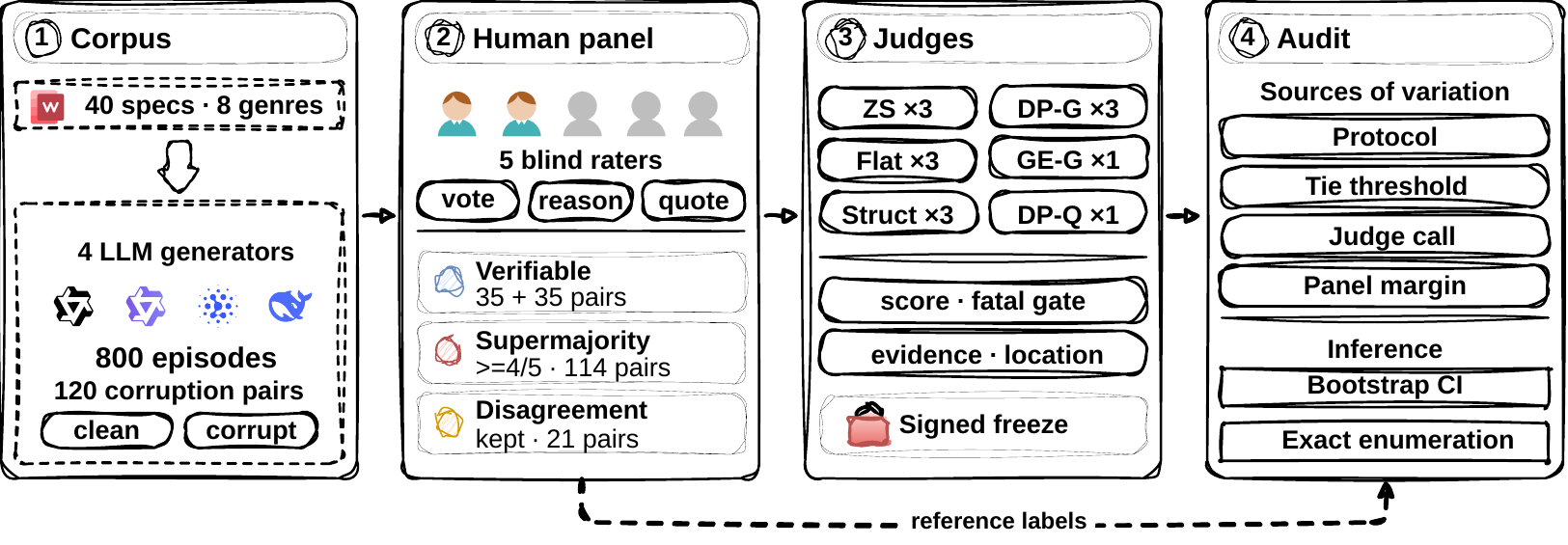}
  \caption{Overview of \method{}. (1) Four LLMs generate 800 episodes from 40 specification packages in eight genres. Each of 120 byte-preserving interventions pairs a corrupted script with its clean source. (2) A fixed five-rater panel labels blinded version pairs and records votes, reasons, and quoted evidence. The records are divided into specification-verifiable constraints, four-of-five panel supermajorities, and within-panel disagreements, each with its own denominator and evidence rule. (3) Six judge protocols return scores, fatal-gate decisions, and localized evidence under a signed freeze. Reference labels are used only during the audit. (4) Two-level bootstrap intervals and exact subpanel enumeration measure variation from protocol, tie calibration, judge call, and panel margin.}
  \label{fig:overview}
\end{figure*}

\section{Related Work}

\paragraph{Drama and long-story evaluation.}
Automatic story metrics agree poorly with human criteria \citep{chhun-etal-2022-human}, which has motivated benchmarks that collect human judgments directly: StoryER combines ranking, rating, and rationales \citep{chen-etal-2022-storyer}, STORIUM provides machine-in-the-loop narratives with human judgments \citep{akoury-etal-2020-storium}, and OpenMEVA adds perturbation tests \citep{guan-etal-2021-openmeva}. Drama-specific work evaluates script continuation along domain dimensions \citep{ma2025dramabench} and studies professional interaction with generated screenplays \citep{mirowski2023dramatron}. For longer narratives, the recurring problems are outline-conditioned coherence \citep{yang-etal-2023-doc,rashkin-etal-2020-plotmachines}, book-length criteria \citep{yang-jin-2025-matters}, preference prediction \citep{fein-etal-2026-litbench}, and subfeature aggregation \citep{feng-etal-2026-howtobench}; surveys name evaluation itself as a persistent bottleneck \citep{alabdulkarim-etal-2021-automatic}. \method{} complements this work by applying separate evidence rules and denominators to constraints, panel supermajorities, and within-panel disagreements.

\paragraph{LLM judges and rubric adaptation.}
G-Eval uses criterion-guided reasoning and form filling \citep{liu-etal-2023-g}, MT-Bench and Chatbot Arena combine model judgments with pairwise human preferences \citep{zheng2023judging,pmlr-v235-chiang24b}, Prometheus trains rubric-following models and extends to direct and pairwise evaluation \citep{kim2024prometheus,kim-etal-2024-prometheus}, and LLM-Rubric calibrates multidimensional judgments against individual raters \citep{hashemi-etal-2024-llm}. Outcomes vary with model identity, response position, response style, and evaluation protocol \citep{wang-etal-2024-large-language-models-fair,chen-etal-2024-humans,shi-etal-2025-judging}; cross-task studies find systematic LLM--human differences \citep{bavaresco-etal-2025-llms}, and stress tests find sensitivity to perturbations \citep{jiang-etal-2026-codejudgebench}. Our experiments therefore include identical-input retests and zero-adaptation evaluations.

Rubric adaptation ranges from generic prompt search \citep{honovich-etal-2023-instruction,zhou2023ape} to fixed criteria with evidence checks \citep{hong-etal-2026-rulers}, model-generated rubrics \citep{siro-etal-2026-learning}, principles or criteria inferred from preferences and comments \citep{findeis2025inverse,xie2025autorubric,yoshida2026feedback}, online criterion elicitation \citep{rezaei2025online}, and human-seeded updates \citep{lin2026growloop}. Our diagnostic evaluates preference agreement together with the preservation of scene-level, episode-level, knowledge-state, and cross-episode constraints.

\paragraph{Learning from disagreement.}
Subjective annotations reflect ambiguity, rater perspective, and value pluralism \citep{uma2021disagreement,davani-etal-2022-dealing,plank-2022-problem}. Perspectivist research examines whether a single aggregate label adequately represents the annotator population \citep{fleisig-etal-2024-perspectivist}. Prescriptive and descriptive annotation encode different objectives \citep{rottger-etal-2022-two}, and jury learning makes panel composition explicit \citep{gordon2022jury}. Language Model Council similarly studies heterogeneous model panels on subjective tasks \citep{zhao-etal-2025-language}. We use supermajority labels from a five-person panel for within-panel preference evaluation and retain below-threshold pairs as within-panel disagreements. These categories describe thresholded outcomes from full vote distributions. Exact enumeration of leave-one-out and three-person subpanels determines which vote margins remain stable under changes in panel composition, while reviewer records are treated as correlated observations.

\section{Problem Setup: Auditing Drama Judges}

Let a generated drama be $X=(E_1,\ldots,E_T)$, where episode $E_t$ contains an ordered sequence of scenes and is generated from specification $Z$. The specification records character personas, relationships, secrets and their initial holders, episode goals, plot nodes, and setup--payoff threads. An evaluation unit $u$ is a scene ($S$), an episode ($E$), or the full series ($F$). Given $(Z,u)$, the evaluator returns
\begin{equation}
J_R(Z,u)=(q,\, f,\, e,\, \ell),
\end{equation}
where $q$ is a quality score under rubric $R$, $f$ is a fatal-gate decision, $e$ is textual evidence, and $\ell$ identifies the scene and episode containing that evidence. The judge must assess the quality of the unit, determine whether it contains a disqualifying defect, and provide localized evidence for both decisions. The only external information available to the judge is $Z$ and the script text. Gold labels, generator identity, and corruption status are excluded from every prompt. We evaluate judge $J_R$ throughout the paper. Success is defined by agreement between the outputs of $J_R$ and the regime-specific reference described below.

\paragraph{Three evaluation regimes.}
In the \verifiable{} regime, correctness is determined from $Z$ and the cross-episode state. A later claim either contradicts an established fact or remains consistent with it. In the \panelsuper{} regime, at least four of five panelists select the same version, and controlled degradations must also pass a manipulation check. In the \withindisagreement{} regime, pairs below the supermajority threshold retain their vote distributions and reasons and are excluded from preference-accuracy calculations. These regimes operationalize the judgments of one fixed five-person panel. Figure~\ref{fig:overview} illustrates the corresponding data paths.

For a panel-supermajority pair $(u^A,u^B)$, an absolute-score judge predicts $\operatorname{sign}(q_A-q_B)$, with absolute score differences of at most $\tau=0.25$ treated as ties. We additionally report results for $\tau\in\{0,.1,.25,.5\}$ and for forced-choice direct judges. For verifiable interventions, successful detection requires a fatal-gate prediction for the corrupted version and no fatal-gate prediction for the matched clean source. Full localization credit further requires identification of the affected episode or scene.

\section{The DramaGrow Benchmark}

\subsection{Structure-Grounded Rubrics}

The detailed seed defines five fatal-gate categories and 16 quality dimensions spread over the dialogue and scene level (L1), the episode-structure level (L2), and the serial-orchestration level (L3), covering causal or logical impossibility, behavior inconsistent with a character, cross-episode contradiction, knowledge-boundary violation, and malformed script structure (Supplement~K). This release instantiates the two information-state gates, cross-episode contradiction and knowledge-boundary violation, as held-out verifiable tests.

A criterion is represented as $c=(d,s,L,A,D,b)$, where $d$ is the criterion description, $s\in\{S,E,F\}$ is the scope, $L$ lists the required evidence locations, $A$ lists the anchors in $Z$, $D$ specifies cross-unit dependencies, and $b$ defines the score boundaries. Evaluating knowledge-boundary management, for example, requires the secret specification, the disclosure event, and the later line or action that uses the information. The flat control expresses the same leaf-level criteria and evidence instructions in prose; the structured rubric adds explicit fields and requires localized evidence.

The judge evaluates each version independently, produces dimension-level scores and evidence, and applies the fatal-gate cascade. Pairwise preference is derived after both evaluations. A single blinded structured output can therefore support the preference, defect-detection, and localization analyses.

\subsection{Repeated Calls and Balanced Fatal Tests}

The main comparison includes three configurations: a zero-shot judge (ZS), a content-matched flat rubric (Flat), and a structured rubric (Struct). The released artifacts retain the historical identifiers C1, C3, and C4 for these configurations. Run 0 is the original frozen run, and runs 1 and 2 independently repeat every input with the same model, endpoint, rubric, reasoning mode, 24K output limit, and service sampling policy. Each run contains 228 side-level evaluations for \PanelSupermajorityPairs{} preference pairs and 70 side-level evaluations for \ConstraintPairsVTwo{} matched clean--corrupted intervention pairs. Missing calls and calls that fail after all retries are counted as errors in the fixed denominator. We report per-run accuracy, the mean, standard deviation, range, three-call majority vote, all-call agreement, and Fleiss' $\kappa$ to characterize call-to-call variation (Supplement~C).

Each configuration is also evaluated on a balanced set of \ConstraintPairsVTwo{} positive and \ConstraintPairsVTwo{} negative items per run, from which we report positive- and negative-specific agreement, three pairwise values of Cohen's $\kappa$, the mean Light's $\kappa$, PABAK, separate I7 and I9 results, and pooled transition counts across runs.

\paragraph{Diagnostic adaptations.}
Earlier experiments examined three forms of automatic rubric modification: an external DramaBench-style rubric (Ext), a flat rubric with heuristic updates (Flat+HL), and coarse- and fine-grained learned structured rubrics (Struct+HL-c and Struct+HL-f). Supplements~F, J, and~K report these experiments. The fine-grained arm selects its static iteration-0 seed, the offline score-reweighting diagnostic (ReWeight; Supplement~G) fails its preregistered gate with a gain of exactly zero percentage points, and none of the learned arms achieves a statistically adjusted gain on the hidden test set. These systems are therefore retained as diagnostic results and excluded from the benchmark ranking.

\subsection{Leakage Firewall}

Series-level splitting is completed before rubric adaptation. Training labels support rubric updates, development labels determine the stopping point, and test artifacts remain sealed during both stages. Before test execution, a signed manifest fixes the prompts, rubrics, exclusion rules, analysis code, and file hashes. Resumed calls use the same signed rubric and model configuration.

\section{Benchmark Corpus and Human Reference}

\subsection{Scripts and Controlled Interventions}

We create 40 setting packages across eight short-drama genres. Four Chinese LLMs, qwen3.7-max, deepseek-v4-pro, glm-4.7, and qwen3.6-flash, independently generate five episodes from each package, producing 160 series versions and 800 episodes. Every generator uses the same prompt and decoding budget: temperature 0.8, top-$p$ 0.95, a 5K output limit, and reasoning disabled. Anonymous labels A--D identify the four generation sources throughout the analysis (Supplement~I). The median episode lengths for A--D are 43, 57, 36, and 54 nonblank lines, respectively. The released pair-level records include these lengths alongside each score difference to support analyses of length sensitivity. The setting packages are stratified by genre into 16 training, 8 development, and 16 hidden-test series groups.

The benchmark contains 120 source--corruption pairs (Supplement~A). Sixty pairs instantiate verifiable defects: knowledge leakage (I7) and cross-episode continuity contradiction (I9). The other 60 instantiate preference-oriented degradations involving dialogue naturalness, character voice, subtext, exposition, nonfunctional scenes, pacing, payoff, and related properties (I1--I6 and I8). Mechanical checks verify that each intended edit is present and that all non-target context is preserved byte for byte. The constraint interventions use explicit anchors from the story specification. Preference interventions enter the accuracy calculation only after the panel confirms the intended loss of quality.

\begin{table}[t]
\centering
\small
\setlength{\tabcolsep}{4pt}
\begin{tabular}{lr}
\toprule
\textbf{Item} & \textbf{Count / value} \\
\midrule
Genres / setting packages & 8 / 40 \\
Generator sources / episodes & 4 / 800 \\
Controlled interventions & 120 \\
Initial annotated pairs / records & 96 / 480 \\
Extension pairs (S/E/F) & \ExtensionAllPairs{} (60/65/10) \\
Extension panel-supermajority / disagreement & \PanelSupermajorityPairs{} / \WithinPanelDisagreementPairs{} \\
Extension 5/5 / 4/5 / 3/5 pairs & \FiveFivePairs{} / \FourFivePairs{} / \ThreeFivePairs{} \\
Fleiss' $\kappa$ & 0.558 \\
Valid / annotated interventions & 27 / 30 \\
Verifiable corrupted / matched clean & 35 / 35 \\
Exact four-person / three-person panels & 5 / 10 per pair \\
\bottomrule
\end{tabular}
\caption{Frozen corpus and human-annotation ledger. A slash in an item name separates the quantities its value reports, in the same order: ``Genres / setting packages'' at ``8 / 40'' means eight genres and 40 packages. All five raters judge every pair, so the 480 initial and 675 extension records are correlated observations of 96 and \ExtensionAllPairs{} pair units, not independent samples.}
\label{tab:data}
\end{table}

\subsection{Five-Rater Annotation}

The annotation set contains 66 same-location pairs sampled across the six unordered generator combinations and 30 blinded preference-intervention pairs. Each rater evaluates 24 scene-level, 48 episode-level, and 24 full-series pairs. A/B order is randomized, model identity and intervention status are hidden, and raters may select ``hard to distinguish.'' For each pair, raters record a preference, confidence, one primary reason, the affected structural level, one quoted evidence span, and an optional fatal issue (Supplement~B).

All five annotators are non-authors, and two have professional experience in drama. Each of the 96 initial pair units is evaluated by five annotators, producing 480 correlated pair--annotator records with complete reasons and evidence spans. The total recorded annotation time is 2,394 minutes, or 39.9 annotator-hours. Fifty-four pairs receive 5/5 agreement, 20 receive 4/5 agreement, 21 receive 3/5 agreement, and one produces a 2/2/1 split. Following established agreement analyses \citep{cohen1960coefficient,fleiss1971measuring,artstein-poesio-2008-survey}, the three-state Fleiss' $\kappa$ is 0.558. All raters exceed the preregistered intervention-attention threshold of 75\%, and 27 of 30 annotated interventions pass the majority manipulation-validity check. Combining the four-of-five rule with this validity check produces 72 initial reference pairs: 32 for training, 15 for development, and 25 for private testing.

Before the labels are revealed, we freeze a nonoverlapping extension of \ExtensionAllPairs{} cross-generator pairs from the 16 test series: 60 scene-level, 65 episode-level, and 10 full-series comparisons. Each position appears once, and quotas are fixed for both generator pairs and series. The same five annotators independently label every pair under blinded conditions. Of these pairs, \PanelSupermajorityPairs{} receive at least four of five votes for one version, while \WithinPanelDisagreementPairs{} remain in the within-panel-disagreement set. The resulting 675 human records support exact composition analyses of all five leave-one-rater-out panels and all ten three-person panels for each pair, with R4--R5 the frozen subgroup with professional drama experience. The panel has three roles: it defines preference references for supermajority pairs, identifies pairs retained as disagreements, and contributes to the rationale-reception audit together with a separate out-of-panel cohort (Supplement~H).

\begin{figure}[t]
  \centering
  \includegraphics[width=\linewidth]{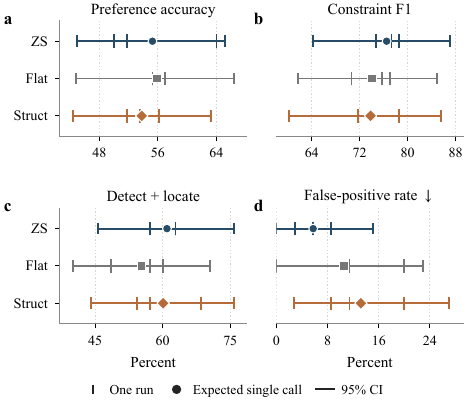}
  \caption{Three independent calls for each core configuration. Each row is one configuration: the three ticks are what run0, run1, and run2 actually returned, the filled marker is the expected single-call estimate, and the capped line is its 95\% two-level bootstrap interval over series and judge calls. Panels are scaled to their own data, so note the differing axes. Denominators are 114 panel-supermajority pairs and 35 I7/I9 interventions with matched clean sources. Reading down a panel: every interval covers every other configuration's estimate, and in (a) the ZS ticks straddle every Flat and Struct run.}
  \label{fig:primary-metrics}
\end{figure}

\begin{figure*}[t]
  \centering
  \includegraphics[width=\textwidth]{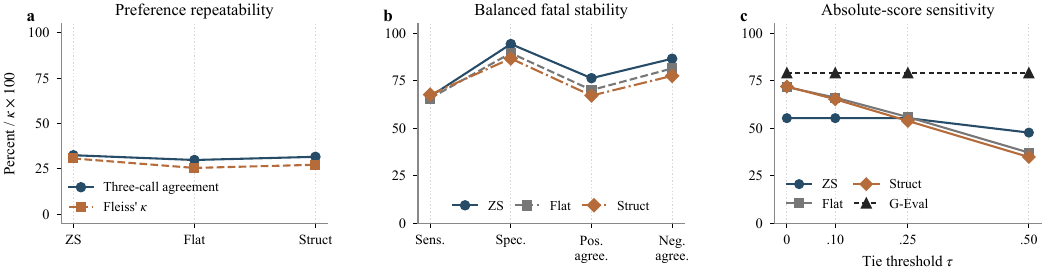}
  \caption{Judge-call stability and score calibration. (a) Three-call preference agreement and Fleiss' $\kappa$. (b) Balanced fatal sensitivity, specificity, and class-specific agreement. (c) Accuracy under absolute-score tie thresholds. Absolute-score accuracy varies with the tie threshold; forced-choice DirectPair has no tie parameter.}
  \label{fig:stability-transfer}
\end{figure*}

\begin{figure}[t]
  \centering
  \includegraphics[width=\linewidth]{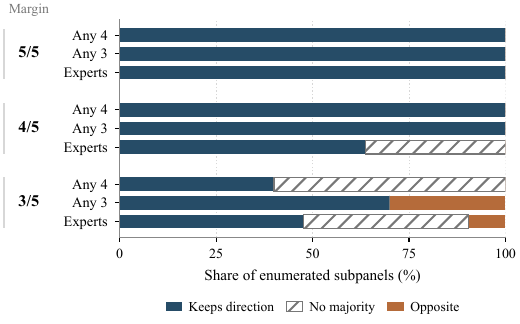}
  \caption{What the four-of-five threshold guarantees, from the existing 675 human records. Every subpanel of every pair is enumerated, so no bar carries sampling error. Each bar reports how one family of subpanels relates to the direction the full panel chose: all five leave-one-rater-out panels, all ten three-person panels, and the two drama-expert raters. A four-person panel can deadlock but can never out-vote the majority, and a three-person panel always decides, so the failure modes differ by family. Four-of-five is the smallest margin no subpanel can overturn; only the professional dyad, which the threshold does not constrain, splits earlier. No machine annotation is used.}
  \label{fig:panel-robustness}
\end{figure}

\section{Experiments}

\paragraph{Research questions.}
\textbf{RQ1:} does ZS, Flat, or Struct hold a reproducible advantage once judge-call variation is included? \textbf{RQ2:} are fatal-gate decisions repeatable on a balanced set of corrupted and clean scripts? \textbf{RQ3:} how much of each vote margin's direction survives exact subpanel enumeration? \textbf{RQ4:} how do DirectPair, G-Eval-style, and open-weight judges compare on the common evaluation sets, and how sensitive are the absolute-score systems to $\tau$?

\begin{table}[t]
\centering
\small
\setlength{\tabcolsep}{3.1pt}
\begin{tabular}{clccc}
\toprule
\textbf{ID} & \textbf{Configuration} & \textbf{Output} & \textbf{Runs} & $\tau$ \\
\midrule
ZS & Zero-shot absolute judge & score+fatal & 3 & yes \\
Flat & Flat static rubric & score+fatal & 3 & yes \\
Struct & Structured static rubric & score+fatal+loc. & 3 & yes \\
DP-G & DirectPair--GLM & forced pair+fatal & 3 & no \\
GE-G & G-Eval--GLM & form score+fatal & 1 & yes \\
DP-Q & DirectPair--Qwen-30B & forced pair+fatal & 1 & no \\
\bottomrule
\end{tabular}
\caption{Main configurations and full-denominator baselines. The released artifacts index ZS, Flat, and Struct as C1, C3, and C4.}
\label{tab:configs}
\end{table}

\paragraph{Configurations and models.}
Table~\ref{tab:configs} lists the three core absolute-score systems and the three full-denominator baselines. Each comparison isolates one design factor: ZS against Flat isolates whether a rubric is provided, Flat against Struct how the same criteria are represented, DP-G the comparison protocol on the same model, and DP-Q the execution model. ZS, Flat, and Struct use GLM-5.2 with reasoning enabled, temperature 1.0, top-$p$ 0.95, and a 24K output limit. DP-G returns a forced anonymous A/B choice with the fatal category and location for each version, GE-G uses fixed criteria, explicit evaluation steps, and a 1--5 form, and DP-Q applies the direct protocol with the open-weight Qwen3-30B-A3B-Instruct-2507 model \citep{qwen3_30b_model_card}. All systems use a deterministic, reversible A/B shuffle and identical script context; gold labels and clean or corrupted status are excluded from every prompt. Each result record includes the task, run, prompt or rubric hash, model, endpoint, and anonymous mapping, and API failures remain errors in the fixed denominator. The earlier ToW and RULERS experiments cover 25 preference pairs and are retained as pilots outside the 114-pair comparison (Supplement~J).

\paragraph{Metrics and inference.}
Preference accuracy is the proportion of predictions that agree with the panel-supermajority reference across \PanelSupermajorityPairs{} extension pairs from \RepeatSeries{} series; we report this raw agreement rate without estimating a Bradley--Terry ranking \citep{bradley1952rank}. On the verifiable track of \ConstraintPairsVTwo{} I7/I9 corruptions and their matched clean sources, sensitivity measures defect detection and specificity measures correct rejection on clean scripts, intervention-macro F1 summarizes that balance within each intervention type, and Detect+Locate additionally requires the evidence to identify the correct episode. The \WithinPanelDisagreementPairs{} below-threshold pairs retain their votes and reasons and are excluded from accuracy calculations.

Confidence intervals are estimated from 10,000 two-level bootstrap samples. Each sample first resamples series and then selects one of the three judge calls for each sampled item \citep{efron1979bootstrap}. The contrasts ZS--Flat, ZS--Struct, and Flat--Struct form one Holm correction family \citep{holm1979simple}. Because low-powered NLP comparisons can produce unstable rankings \citep{dror-etal-2018-hitchhikers,card-etal-2020-little}, we also report results for each run and for the three-call majority ensemble. These additional results preserve information about run-to-run variation beyond the confidence intervals. Absolute-score systems are evaluated at $\tau\in\{0,.1,.25,.5\}$, with the corresponding tie rates and score-difference distributions. Direct systems use forced choice, and exact subpanel enumeration introduces no Monte Carlo error.

\subsection{Main Comparisons}

The repeated-call intervals of the three absolute-score rubrics overlap (RQ1). Figure~\ref{fig:primary-metrics} reports all three calls. Mean panel-supermajority accuracy is \COnePrefMean\% for ZS, \CThreePrefMean\% for Flat, and \CFourPrefMean\% for Struct. The corresponding run ranges are [\COnePrefLow, \COnePrefHigh], [\CThreePrefLow, \CThreePrefHigh], and [\CFourPrefLow, \CFourPrefHigh] percent. The expected single-call contrasts are ZS--Flat $=\COneCThreeDelta$ percentage points (95\% CI [\COneCThreeCILow, \COneCThreeCIHigh], Holm-adjusted $p=\COneCThreeHolm$), ZS--Struct $=\COneCFourDelta$ percentage points (95\% CI [\COneCFourCILow, \COneCFourCIHigh], $p=\COneCFourHolm$), and Flat--Struct $=\CThreeCFourDelta$ percentage points (95\% CI [\CThreeCFourCILow, \CThreeCFourCIHigh], $p=\CThreeCFourHolm$). The ZS results range from \COnePrefLow\% to \COnePrefHigh\%, spanning values below and above every run of Flat and Struct. We therefore characterize each configuration by its repeated-call distribution.

The balanced fatal track yields similar sensitivity across rubric variants but different false-positive rates on clean scripts (RQ2). Mean sensitivity and specificity are \COneSensitivity\% and \COneSpecificity\% for ZS, \CThreeSensitivity\% and \CThreeSpecificity\% for Flat, and \CFourSensitivity\% and \CFourSpecificity\% for Struct. Mean Detect+Locate is \COneDetectLocate\%, \CThreeDetectLocate\%, and \CFourDetectLocate\%, respectively. The separation sits on the clean sources: at similar detection rates, the false-positive rate is \COneFPRVTwo\% for ZS against \CFourFPRVTwo\% for Struct, so the added structure buys no extra detection here but alarms more often on clean scripts.

\paragraph{Unified strong baselines.}
Evaluation protocol produces the largest measured difference in the full-denominator comparison (RQ4). DirectPair--GLM achieves \DirectGLMPreference\% mean preference accuracy, \DirectGLMConstraintFOne\% constraint F1, and \DirectGLMDetectLocate\% Detect+Locate. Each of its three runs achieves \DirectGLMRunZeroPreference\%, which exceeds the highest single-run accuracy among the absolute-score systems. Three-call agreement at the item level is \DirectGLMAgreement\% ($\kappa=\DirectGLMKappa$), indicating that identical aggregate accuracy can include different item-level decisions. G-Eval--GLM achieves \GEvalGLMPreference\% preference accuracy, \GEvalGLMConstraintFOne\% constraint F1, and \GEvalGLMDetectLocate\% Detect+Locate. The corresponding results for DirectPair--Qwen are \DirectQwenPreference\%, \DirectQwenConstraintFOne\%, and \DirectQwenDetectLocate\%. All systems use the same sets of 114 preference pairs and 35 constraint interventions (Supplement~D), which excludes denominator size as an explanation for the differences. DP-G and the absolute-score systems use the same GLM, enabling a within-model comparison of protocol bundles. DP-Q evaluates transfer of the direct protocol to an open-weight model.

\begin{table}[t]
\centering
\scriptsize
\setlength{\tabcolsep}{2.4pt}
\begin{tabular}{lrcrrrr}
\toprule
\textbf{System} & \textbf{Pref.} & \textbf{Run range} & \textbf{F1} & \textbf{D+L} & \textbf{FPR$\downarrow$} & \textbf{Calls} \\
\midrule
ZS & \COnePrefMean & [\COnePrefLow, \COnePrefHigh] & \COneConstraintFOne & \secondbest{\COneDetectLocate} & \COneFPRVTwo & 3 \\
Flat & \CThreePrefMean & [\CThreePrefLow, \CThreePrefHigh] & \CThreeConstraintFOne & \CThreeDetectLocate & \CThreeFPRVTwo & 3 \\
Struct & \CFourPrefMean & [\CFourPrefLow, \CFourPrefHigh] & \CFourConstraintFOne & \CFourDetectLocate & \CFourFPRVTwo & 3 \\
\midrule
\rowcolor{rowhl}
DP-G & \best{\DirectGLMPreference} & [\DirectGLMRunZeroPreference, \DirectGLMRunTwoPreference] & \best{\DirectGLMConstraintFOne} & \best{\DirectGLMDetectLocate} & \best{\DirectGLMFPR} & 3 \\
GE-G & \GEvalGLMPreference & -- & \GEvalGLMConstraintFOne & \GEvalGLMDetectLocate & \secondbest{\GEvalGLMFPR} & 1 \\
DP-Q & \secondbest{\DirectQwenPreference} & -- & \secondbest{\DirectQwenConstraintFOne} & \DirectQwenDetectLocate & \DirectQwenFPR & 1 \\
\bottomrule
\end{tabular}
\caption{Unified results on the common sets of 114 preference pairs and 35 constraint interventions (percent). \best{Bold} indicates the best value in each column, and \secondbest{underlining} indicates the second-best value. The DirectPair--GLM row is shaded because it has the best value in four of the five performance columns. Three-call entries are means, with the minimum and maximum preference accuracies across runs shown as ranges. Single-call baselines are evaluated in one frozen run. Preference accuracy is identical across the three runs of DP-G.}
\label{tab:unified-results}
\end{table}

\subsection{Decision Repeatability}

Item-level decisions are substantially less stable than the aggregate percentages suggest. Figure~\ref{fig:stability-transfer} reports preference repeatability, balanced fatal-gate reliability, and calibration. Three-call preference agreement is \COnePrefAgreement\% for ZS, \CThreePrefAgreement\% for Flat, and \CFourPrefAgreement\% for Struct, with corresponding Fleiss' $\kappa$ values of \COnePrefKappa, \CThreePrefKappa, and \CFourPrefKappa. Fatal-gate decisions are more repeatable than preference decisions. Positive-specific agreement is \COnePositiveAgreement\%, \CThreePositiveAgreement\%, and \CFourPositiveAgreement\%, and negative-specific agreement is \COneNegativeAgreement\%, \CThreeNegativeAgreement\%, and \CFourNegativeAgreement\%, respectively. The corresponding mean values of Light's $\kappa$ are \COneLightKappa, \CThreeLightKappa, and \CFourLightKappa. The balanced design estimates reproducibility separately for positive and negative items.

Changing $\tau$ affects both accuracy and tie rate for ZS, Flat, Struct, and G-Eval. Figure~\ref{fig:stability-transfer} reports the complete threshold curves. DirectPair uses forced choice and therefore has no tie threshold. The Qwen baseline changes both the execution model and the scoring protocol and serves as an open-weight portability check.

\section{Analysis, Scope, and Ethics}

\paragraph{Why four-of-five is the threshold.}
To answer RQ3, Figure~\ref{fig:panel-robustness} uses the 675 completed human records to evaluate the four-of-five rule (Supplement~E). By construction, a 5/5 or 4/5 majority retains its direction in every leave-one-rater-out and three-person subpanel, and exact enumeration confirms 100\% retention for all \FiveFivePairs{} pairs at 5/5 and all \FourFivePairs{} at 4/5. At 3/5, leave-one-out panels retain the full-panel direction only \ThreeFiveLeaveRetention\% of the time and go unresolved \ThreeFiveLeaveUnresolved\%, while three-person panels reverse it \ThreeFiveTripleReversal\% of the time. Four-of-five is therefore the smallest vote margin invariant to the removal of any one rater, which is why the 21 pairs below it remain in the disagreement set rather than receiving a preference label.

\paragraph{What the structured rubric isolates.}
The clearest observed effect of the structured rubric concerns the reception of rationales, not the accuracy of decisions. Struct bundles explicit criterion fields with mandatory localization, specification anchors, and a fixed output format, so the Flat--Struct contrast estimates their combined effect and the released artifacts carry the per-field attribution. In the out-of-panel audit (Supplement~H), Struct receives \IndCFourReasonCount{} majority decisions for reasonableness, \IndCFourEvidenceCount{} for evidence support, and \IndCFourCoverageCount{} for reason coverage. It also produces the longest outputs, so we report rationale reception separately from decision accuracy.

\paragraph{Judge and protocol dependence.}
The systems in Table~\ref{tab:unified-results} differ in scoring protocol, input format, and execution model, and each difference carries a deployment meaning. DirectPair sees both versions, so its margin describes screening in which a reference draft is available; ZS reads one script at a time and describes single-script screening. The two settings are not interchangeable, and the DP-G margin does not transfer to the single-script case. G-Eval combines \GEvalGLMSensitivity\% sensitivity with \GEvalGLMSpecificity\% specificity, a conservative fatal-gate policy that rejects few clean scripts and misses most defects. DP-Q retains most of the direct protocol's preference accuracy on an open-weight model, but not its localization or clean-side calibration.

Tie calibration moves the absolute-score systems as much as any rubric choice does: Flat and Struct fall from \CThreeTauZeroAccuracy\% and \CFourTauZeroAccuracy\% at $\tau=0$ to \CThreeTauPointFiveAccuracy\% and \CFourTauPointFiveAccuracy\% at $\tau=.5$, and their order reverses along the way.

\paragraph{Limitations.}
The conclusions of this audit are limited to the frozen evaluation sets in several respects. First, the corpus consists of synthetic Chinese five-episode vertical short-form dramas from eight genres and \RepeatSeries{} held-out series clusters. Generalization to other languages, narrative lengths, and production settings requires separate evaluation. Second, the human reference is derived from one fixed five-person panel. A panel with a different composition may define different boundaries between supermajority preference and disagreement. Third, the verifiable experiment covers two information-state gates, knowledge-boundary violation and cross-episode continuity violation, using matched clean sources. The other three gates are specified but have no test instances in this release. Repeatability estimates are based on three calls, each HL seed has one optimizer trajectory, and the cross-judge analysis uses one GLM-family judge pipeline. Future releases will evaluate the remaining fatal gates, longer serial narratives, additional genres and languages, and additional human panels.

\paragraph{Ethics and data governance.}
All scripts in the benchmark are synthetic. The five non-author annotators and the independent audit reviewers participated voluntarily under disclosed protocols, with compensation or authorship arranged before participation. The subpanel analysis reuses their completed records, and every label in it is human. Released identifiers are anonymized, attribution-session records remain private, and raw labels are immutable.

\paragraph{Release and reproducibility.}
The release separates four provenance layers: scripts and specification packages; controlled source--corruption pairs; anonymized human preferences, reasons, and evidence spans; and frozen judge executions. Every prediction is bound to its task, configuration, run, input and rubric hashes, and model identifiers, and denominator records report expected, valid, and failed counts, so a call that exhausts the format-retry policy stays visible rather than disappearing from a denominator. Offline scripts reconstruct every reported number, from panel margins to bootstrap intervals and figure data, without further model calls, and historical pilots and audits are stored in separately named artifact families (Supplement~J). A new judge can therefore be evaluated zero-adaptation on the released sets.

\section{Conclusion}

We introduced \method{}, a regime-aware benchmark and audit protocol for evaluating LLM judges of AI-generated multi-episode drama: 800 episodes, 120 paired interventions, 1{,}155 blinded five-rater records, regime-specific evaluation rules, and offline audit code. On its frozen sets the repeated-call results of the absolute-score rubrics overlap, while DirectPair--GLM leads on the same GLM at every run and tie threshold, so protocol outweighs rubric representation here. The results support four reporting practices: repeat judge calls, preregister and sweep the tie threshold, pair corrupted tests with clean sources, and retain votes and reasons for below-threshold pairs. This audit design should transfer to other long-form creative tasks that mix verifiable errors, majority preferences, and human disagreement.

\FloatBarrier
\clearpage

\bibliography{references}

\clearpage
\setcounter{section}{0}
\setcounter{subsection}{0}
\setcounter{secnumdepth}{2}
\renewcommand{\thesection}{\Alph{section}}
\renewcommand{\thesubsection}{\thesection.\arabic{subsection}}
\twocolumn[\begin{center}\Large\bfseries Supplementary Material\end{center}\vspace{1em}]

\paragraph{System naming.}
This supplement uses the descriptive system names adopted in the main paper. The released artifacts, frozen logs, and manifests retain the historical identifiers. ZS corresponds to C1 (zero-shot judge), Ext to C2 (external DramaBench-style rubric), Flat to C3 (flat static rubric), Struct to C4 (structured static rubric), Flat+HL to C6 (flat rubric with heuristic updates), Struct+HL-c and Struct+HL-f to C7-c and C7-d (coarse- and fine-grained learned structured rubrics), and ReWeight to C8 (offline score-reweighting diagnostic).

\section{Intervention Taxonomy}
\label{supp:interventions}

The 120 controlled interventions support the regime-specific analyses in the benchmark. Story-specification anchors and contradictory text spans make I7 and I9 directly verifiable, and these intervention types constitute the held-out constraint track. Independent human judgments validate I1--I6 and I8 before these interventions enter the panel-supermajority accuracy calculation. All non-target episodes are copied byte for byte from the source.

\begin{center}
\centering
\scriptsize
\setlength{\tabcolsep}{2.5pt}
\begin{tabular}{cp{1.40in}cc}
\toprule
\textbf{ID} & \textbf{Intervention} & \textbf{Level} & \textbf{Regime} \\
\midrule
I1 & Remove episode-end hook & L2 & Panel pref. \\
I2 & Flatten a reversal & L2 & Panel pref. \\
I3 & Delete setup, keep payoff & L2 & Panel pref. \\
I4 & Inject a nonfunctional scene & L2 & Panel pref. \\
I5 & Swap dialogue across roles & L1 & Panel pref. \\
I6 & Remove a relationship trigger & L3 & Panel pref. \\
I7 & Violate a knowledge boundary & L3 & Verif. \\
I8 & Transfer protagonist agency & L3 & Panel pref. \\
I9 & Introduce continuity conflict & L3 & Verif. \\
\bottomrule
\end{tabular}
\captionof{table}{Nine controlled intervention types.}
\label{tab:interventions}
\end{center}

The expected evidence location depends on the intervention type: the modified hook for I1, reveal order for I2, the missing setup and retained payoff for I3, the inserted scene for I4, reassigned dialogue for I5, the missing relationship trigger for I6, the secret specification and violating line for I7, the transferred climactic action for I8, and the two contradictory facts for I9. I7 is expected to trigger fatal gate F4, and I9 is expected to trigger fatal gate F3.

\section{Annotation Protocol and Frozen Extension}
\label{supp:annotation}

Five non-author raters independently annotate the same 96 blinded pairs. A/B order is randomized independently of generator identity, and the annotation cards conceal the provenance of each pair. The 24 scene-level units include the target scene and character specification. The 48 episode-level units include one episode and a three-line summary of the preceding context. The 24 full-series units include both five-episode versions and the corresponding setting package.

For each pair, raters record (1) a preference of A, B, or hard to distinguish; (2) confidence on a three-point scale; (3) one primary reason; (4) the affected level, L1, L2, L3, or multiple levels; (5) one quoted evidence line; and (6) an optional fatal issue with the affected version and location. Raw labels are immutable. Subsequent attribution decisions are stored in a separate provenance layer and do not alter the five independent judgments.

The annotation data contain 96 independent pair units, each evaluated by five raters, for a total of 480 correlated pair--annotator records. Every record includes a reason and an evidence span. All five raters exceed the intervention-attention threshold of 75\%. The overall three-state Fleiss' $\kappa$ is 0.558, and Cohen's $\kappa$ for the two raters with professional drama experience is 0.729. Annotation time is recorded in minutes. The five raters contribute 499, 237, 369, 866, and 423 minutes, respectively, and all raters have equal voting weight.

Before the labels are revealed, a signed manifest fixes \ExtensionAllPairs{} nonoverlapping test pairs: 60 scene-level, 65 episode-level, and 10 full-series comparisons. The extension contains no positions from the original set of 96 pairs. Each new position appears once, each unordered generator combination contributes 22--23 pairs, and each of the 16 test series contributes 8--9 pairs. The original five annotators independently label every pair under blinded conditions. A total of \PanelSupermajorityPairs{} pairs receive at least four of five votes for one version and form the panel-supermajority set. The remaining \WithinPanelDisagreementPairs{} pairs form the within-panel-disagreement set and retain their full vote distributions and reasons.

\section{Three-Call Core Results}
\label{supp:threecall}

The primary comparison includes ZS, Flat, and Struct and combines the original frozen run with two independent full repeats. Each run evaluates \PanelSupermajorityPairs{} preference pairs and \ConstraintPairsVTwo{} corrupted interventions together with their \ConstraintPairsVTwo{} matched clean sources. A missing side-level output or a call that fails after all retries is counted as an error in the fixed denominator.

\begin{center}
\scriptsize
\setlength{\tabcolsep}{3.5pt}
\begin{tabular}{lrrrrr}
\toprule
\textbf{Config.} & \textbf{run0} & \textbf{run1} & \textbf{run2} & \textbf{Mean} & \textbf{Majority call} \\
\midrule
ZS & \COneRunZeroPref & \COneRunOnePref & \COneRunTwoPref & \COnePrefMean & \COnePrefMajority \\
Flat & \CThreeRunZeroPref & \CThreeRunOnePref & \CThreeRunTwoPref & \CThreePrefMean & \CThreePrefMajority \\
Struct & \CFourRunZeroPref & \CFourRunOnePref & \CFourRunTwoPref & \CFourPrefMean & \CFourPrefMajority \\
\bottomrule
\end{tabular}
\captionof{table}{Panel-supermajority preference accuracy (percent) for the common set of 114 pairs.}
\label{tab:v2-preference-runs}
\end{center}

\begin{center}
\scriptsize
\setlength{\tabcolsep}{2.7pt}
\begin{tabular}{llrrrrrr}
\toprule
\textbf{Config.} & \textbf{Run} & \textbf{Sens.} & \textbf{Spec.} & \textbf{Bal. acc.} & \textbf{F1} & \textbf{D+L} & \textbf{FPR} \\
\midrule
ZS & run0 & \COneRunZeroSensitivity & \COneRunZeroSpecificity & \COneRunZeroBalanced & \COneRunZeroFOne & \COneRunZeroLocate & \COneRunZeroFPR \\
   & run1 & \COneRunOneSensitivity & \COneRunOneSpecificity & \COneRunOneBalanced & \COneRunOneFOne & \COneRunOneLocate & \COneRunOneFPR \\
   & run2 & \COneRunTwoSensitivity & \COneRunTwoSpecificity & \COneRunTwoBalanced & \COneRunTwoFOne & \COneRunTwoLocate & \COneRunTwoFPR \\
Flat & run0 & \CThreeRunZeroSensitivity & \CThreeRunZeroSpecificity & \CThreeRunZeroBalanced & \CThreeRunZeroFOne & \CThreeRunZeroLocate & \CThreeRunZeroFPR \\
   & run1 & \CThreeRunOneSensitivity & \CThreeRunOneSpecificity & \CThreeRunOneBalanced & \CThreeRunOneFOne & \CThreeRunOneLocate & \CThreeRunOneFPR \\
   & run2 & \CThreeRunTwoSensitivity & \CThreeRunTwoSpecificity & \CThreeRunTwoBalanced & \CThreeRunTwoFOne & \CThreeRunTwoLocate & \CThreeRunTwoFPR \\
Struct & run0 & \CFourRunZeroSensitivity & \CFourRunZeroSpecificity & \CFourRunZeroBalanced & \CFourRunZeroFOne & \CFourRunZeroLocate & \CFourRunZeroFPR \\
   & run1 & \CFourRunOneSensitivity & \CFourRunOneSpecificity & \CFourRunOneBalanced & \CFourRunOneFOne & \CFourRunOneLocate & \CFourRunOneFPR \\
   & run2 & \CFourRunTwoSensitivity & \CFourRunTwoSpecificity & \CFourRunTwoBalanced & \CFourRunTwoFOne & \CFourRunTwoLocate & \CFourRunTwoFPR \\
\bottomrule
\end{tabular}
\captionof{table}{Balanced fatal-gate results (percent) for 35 positive and 35 negative items in each run. F1 is the macro-average across the I7 and I9 interventions. D+L requires both detection and correct episode localization.}
\label{tab:v2-balanced-fatal-runs}
\end{center}

\begin{center}
\scriptsize
\setlength{\tabcolsep}{3.6pt}
\begin{tabular}{lrrrrrr}
\toprule
\textbf{Config.} & \textbf{Pref. agree} & \textbf{Pref. $\kappa$} & \textbf{Pos. agree} & \textbf{Neg. agree} & \textbf{Light $\kappa$} & \textbf{PABAK} \\
\midrule
ZS & \COnePrefAgreement & \COnePrefKappa & \COnePositiveAgreement & \COneNegativeAgreement & \COneLightKappa & \COnePABAK \\
Flat & \CThreePrefAgreement & \CThreePrefKappa & \CThreePositiveAgreement & \CThreeNegativeAgreement & \CThreeLightKappa & \CThreePABAK \\
Struct & \CFourPrefAgreement & \CFourPrefKappa & \CFourPositiveAgreement & \CFourNegativeAgreement & \CFourLightKappa & \CFourPABAK \\
\bottomrule
\end{tabular}
\captionof{table}{Preference reliability and balanced fatal-gate reliability across three calls. Preference agreement requires all three calls to produce the same decision. Fatal-gate agreement is averaged over the three pairs of runs on the balanced vector of 70 items.}
\label{tab:v2-reliability}
\end{center}

\begin{center}
\scriptsize
\setlength{\tabcolsep}{4.0pt}
\begin{tabular}{llrrrr}
\toprule
\textbf{Config.} & \textbf{Type} & \textbf{Sens.} & \textbf{Spec.} & \textbf{F1} & \textbf{D+L} \\
\midrule
ZS & I7 & \COneISevenSensitivity & \COneISevenSpecificity & \COneISevenFOne & \COneISevenLocate \\
   & I9 & \COneINineSensitivity & \COneINineSpecificity & \COneINineFOne & \COneINineLocate \\
Flat & I7 & \CThreeISevenSensitivity & \CThreeISevenSpecificity & \CThreeISevenFOne & \CThreeISevenLocate \\
   & I9 & \CThreeINineSensitivity & \CThreeINineSpecificity & \CThreeINineFOne & \CThreeINineLocate \\
Struct & I7 & \CFourISevenSensitivity & \CFourISevenSpecificity & \CFourISevenFOne & \CFourISevenLocate \\
   & I9 & \CFourINineSensitivity & \CFourINineSpecificity & \CFourINineFOne & \CFourINineLocate \\
\bottomrule
\end{tabular}
\captionof{table}{Results for I7 knowledge-boundary violations and I9 cross-episode continuity violations, averaged across the three runs (percent).}
\label{tab:v2-i7-i9}
\end{center}

\section{Full-Denominator Strong Baselines}
\label{supp:baselines}

All three baselines use the same sets of 114 preference pairs and 35 constraint interventions. DirectPair--GLM is evaluated in three independent runs. G-Eval--GLM and DirectPair--Qwen3-30B-A3B-Instruct-2507 are each evaluated in one frozen run. The deterministic anonymous A/B mappings can be reversed using the private provenance records, and failed calls remain errors in the denominator.

\begin{center}
\scriptsize
\setlength{\tabcolsep}{3.7pt}
\begin{tabular}{lrrrr}
\toprule
\textbf{Baseline} & \textbf{Preference} & \textbf{Constraint F1} & \textbf{D+L} & \textbf{FPR} \\
\midrule
\rowcolor{rowhl}
DirectPair--GLM & \best{\DirectGLMPreference} & \best{\DirectGLMConstraintFOne} & \best{\DirectGLMDetectLocate} & \best{\DirectGLMFPR} \\
G-Eval--GLM & \GEvalGLMPreference & \GEvalGLMConstraintFOne & \GEvalGLMDetectLocate & \secondbest{\GEvalGLMFPR} \\
DirectPair--Qwen & \secondbest{\DirectQwenPreference} & \secondbest{\DirectQwenConstraintFOne} & \secondbest{\DirectQwenDetectLocate} & \DirectQwenFPR \\
\bottomrule
\end{tabular}
\captionof{table}{Unified baseline estimates (percent). \best{Bold} indicates the best value in each column, and \secondbest{underlining} indicates the second-best value. The DirectPair--GLM row is shaded because it has the best value in all four columns. Entries for DirectPair--GLM are means over three runs; the other entries are from one frozen run.}
\label{tab:v2-baselines}
\end{center}

The preference accuracies of DirectPair--GLM in runs 0, 1, and 2 are \DirectGLMRunZeroPreference\%, \DirectGLMRunOnePreference\%, and \DirectGLMRunTwoPreference\%, respectively. The corresponding constraint F1 values are \DirectGLMRunZeroFOne\%, \DirectGLMRunOneFOne\%, and \DirectGLMRunTwoFOne\%. Three-call preference agreement is \DirectGLMAgreement\%, with Fleiss' $\kappa=\DirectGLMKappa$.

\begin{figure}[t]
\centering
\begin{tikzpicture}
\begin{axis}[
  name=accplot,
  width=0.53\linewidth, height=1.72in,
  xlabel={Tie threshold $\tau$}, ylabel={Accuracy (\%)},
  xtick={0,0.1,0.25,0.5}, xticklabels={0,.1,.25,.5},
  ymin=0, ymax=100,
  tick label style={font=\scriptsize},
  label style={font=\scriptsize},
  legend style={font=\tiny, at={(0.03,0.05)}, anchor=south west, draw=none, fill=none},
  legend columns=2,
]
\addplot+[mark=*] coordinates {(0,\COneTauZeroAccuracy) (0.1,\COneTauPointOneAccuracy) (0.25,\COneTauPointTwoFiveAccuracy) (0.5,\COneTauPointFiveAccuracy)};
\addplot+[mark=square*] coordinates {(0,\CThreeTauZeroAccuracy) (0.1,\CThreeTauPointOneAccuracy) (0.25,\CThreeTauPointTwoFiveAccuracy) (0.5,\CThreeTauPointFiveAccuracy)};
\addplot+[mark=diamond*] coordinates {(0,\CFourTauZeroAccuracy) (0.1,\CFourTauPointOneAccuracy) (0.25,\CFourTauPointTwoFiveAccuracy) (0.5,\CFourTauPointFiveAccuracy)};
\addplot+[mark=triangle*, dashed] coordinates {(0,\GEvalTauZeroAccuracy) (0.1,\GEvalTauPointOneAccuracy) (0.25,\GEvalTauPointTwoFiveAccuracy) (0.5,\GEvalTauPointFiveAccuracy)};
\legend{ZS,Flat,Struct,G-Eval}
\end{axis}
\begin{axis}[
  at={(accplot.outer east)}, anchor=outer west, xshift=2pt,
  width=0.53\linewidth, height=1.72in,
  xlabel={Tie threshold $\tau$}, ylabel={Tie rate (\%)},
  xtick={0,0.1,0.25,0.5}, xticklabels={0,.1,.25,.5},
  ymin=0, ymax=100,
  tick label style={font=\scriptsize},
  label style={font=\scriptsize},
  legend style={font=\tiny, at={(0.03,0.97)}, anchor=north west, draw=none, fill=none},
  legend columns=2,
]
\addplot+[mark=*] coordinates {(0,\COneTauZeroTie) (0.1,\COneTauPointOneTie) (0.25,\COneTauPointTwoFiveTie) (0.5,\COneTauPointFiveTie)};
\addplot+[mark=square*] coordinates {(0,\CThreeTauZeroTie) (0.1,\CThreeTauPointOneTie) (0.25,\CThreeTauPointTwoFiveTie) (0.5,\CThreeTauPointFiveTie)};
\addplot+[mark=diamond*] coordinates {(0,\CFourTauZeroTie) (0.1,\CFourTauPointOneTie) (0.25,\CFourTauPointTwoFiveTie) (0.5,\CFourTauPointFiveTie)};
\addplot+[mark=triangle*, dashed] coordinates {(0,\GEvalTauZeroTie) (0.1,\GEvalTauPointOneTie) (0.25,\GEvalTauPointTwoFiveTie) (0.5,\GEvalTauPointFiveTie)};
\legend{ZS,Flat,Struct,G-Eval}
\end{axis}
\end{tikzpicture}
\caption{Tie-threshold sensitivity of the absolute-score systems: accuracy (left) and tie rate (right) at $\tau\in\{0,.1,.25,.5\}$. Every coordinate is a frozen repeatability-v2 macro; forced-choice DirectPair has no $\tau$ parameter.}
\label{fig:s-tau}
\end{figure}
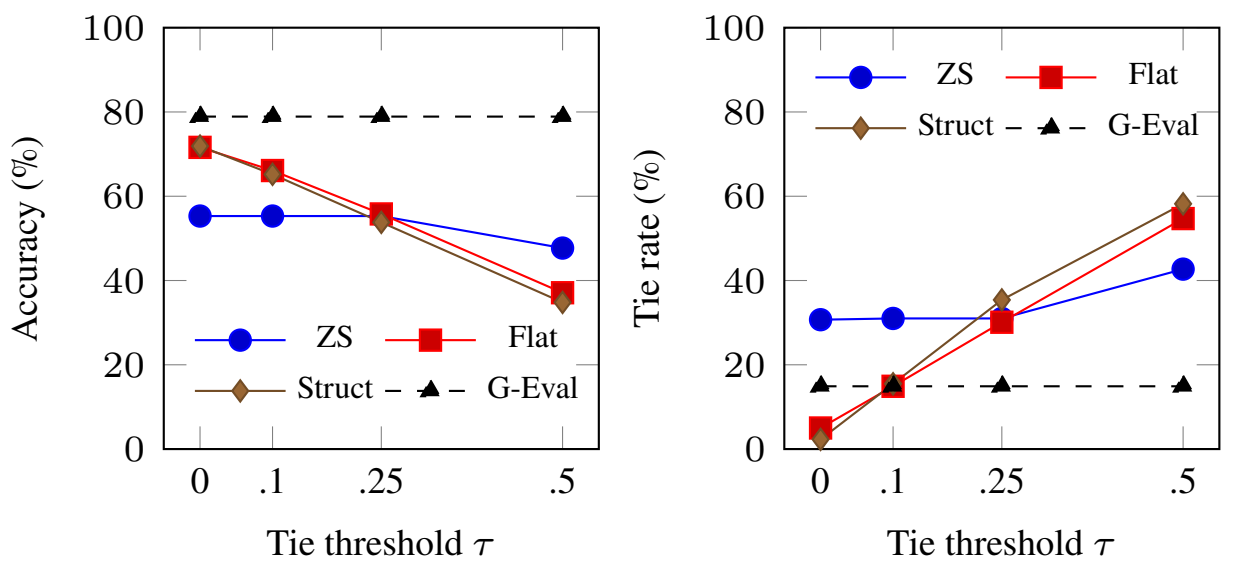

Figure~\ref{fig:s-tau} shows sensitivity to the tie threshold for the absolute-score systems. Both accuracy and tie rate change with $\tau$. At the frozen threshold of $\tau=0.25$, approximately one-third of the ZS, Flat, and Struct decisions are ties; at $\tau=0.5$, more than half are ties. The coarse 1--5 output scale of G-Eval produces the same values at every evaluated threshold. DirectPair uses forced choice and has no $\tau$ parameter.

\section{Exact Subpanel Enumeration by Vote Margin}
\label{supp:subpanel}

The 675 completed human records correspond to \FiveFivePairs{} pairs with 5/5 agreement, \FourFivePairs{} pairs with 4/5 agreement, and \ThreeFivePairs{} pairs with 3/5 agreement. For every pair, we enumerate all five leave-one-rater-out panels and all ten three-person panels. No subpanels are sampled, so the resulting counts contain no Monte Carlo error.

A vote margin of 5/5 or 4/5 retains its direction in every enumerated subpanel by construction. Exact enumeration confirms 100\% retention and no unresolved outcomes for both subpanel families. Pairs with 3/5 agreement do not satisfy this property. Leave-one-out panels retain the full-panel direction \ThreeFiveLeaveRetention\% of the time and produce an unresolved result \ThreeFiveLeaveUnresolved\% of the time. Three-person panels retain the direction \ThreeFiveTripleRetention\% of the time and reverse it \ThreeFiveTripleReversal\% of the time. Figure~4 of the main paper presents these values, and the release includes the pair-level enumeration records. R1--R3 form the fixed nonprofessional subgroup, and R4--R5 form the subgroup with professional drama experience.

\section{Historical Single-Call Diagnostics}
\label{supp:historical}

This section reports the single-call estimates obtained before the three-call analysis. The primary system comparison is based on the repeated-call results in Section~\ref{supp:threecall}. Panel-supermajority accuracy in this section uses the 114 extension pairs, and the constraint metrics use the 35 I7/I9 interventions. Section~\ref{supp:pilots} reports the corresponding original 25-pair pilot estimates once for the primary GLM judge. The historical analysis resolves a score difference exactly equal to $\tau$ as a decision, whereas the repeated-call analysis in Section~\ref{supp:threecall} treats it as a tie; one Flat pair in the original run lies exactly on this boundary, which accounts for the one-pair difference between the Flat entries in Table~\ref{tab:complete-primary} and Table~\ref{tab:v2-preference-runs}.

\begin{center}
\centering
\scriptsize
\setlength{\tabcolsep}{2.5pt}
\begin{tabular}{lcccc}
\toprule
\textbf{Config.} & \textbf{Panel-super.} & \textbf{Constraint F1} & \textbf{D+L} & \textbf{FPR} \\
\midrule
ZS & \COneConsensus & \COneConstraint & \COneLocation & \COneFPR \\
Ext & \CTwoConsensus & \CTwoConstraint & \CTwoLocation & \CTwoFPR \\
Flat & \CThreeConsensus & \CThreeConstraint & \CThreeLocation & \CThreeFPR \\
Struct & \CFourConsensus & \CFourConstraint & \CFourLocation & \CFourFPR \\
Flat+HL & \CSixConsensus & \CSixConstraint & \CSixLocation & \CSixFPR \\
Struct+HL-c & \CSevenCConsensus & \CSevenCConstraint & \CSevenCLocation & \CSevenCFPR \\
\bottomrule
\end{tabular}
\captionof{table}{Historical single-call results for the primary judge (percent). Brackets contain 95\% series-clustered intervals. D+L denotes joint detection--localization.}
\label{tab:complete-primary}
\end{center}

The four preregistered expanded-test contrasts are Struct--Flat, Flat+HL--Flat, Struct+HL-f--Flat+HL, and Struct+HL-c--Struct. The corresponding changes in accuracy are \DeltaStructureConsensus, \DeltaFlatHLConsensus, \DeltaDetailedConsensus, and \DeltaCoarseConsensus{} percentage points. Every clustered interval includes zero, and all four Holm-adjusted $p$-values are 1.0. The single-call results show no statistically adjusted gain on the hidden test set for either learned or restructured rubrics. We retain these results as provenance for the diagnostic experiments.

\section{Frozen Zero-Cost ReWeight Diagnostic}
\label{supp:c8}

ReWeight is a fully offline aggregation diagnostic with a two-stage freeze. During the freeze stage, it can access the 32 training pairs, 15 development pairs, and side-level outputs from Struct. Extension labels are excluded from all accessible data paths and tests. Fatal gates, score caps, evidence, localization, and all 35 constraint records remain byte-equivalent to the corresponding Struct outputs. The only optimized parameters are nonnegative L1/L2/L3 weights on a grid with step size 0.05 and within an L1 distance of 0.50 from the original weights of Struct. Pair decisions are binary. A seeded hash resolves only cases with an exact score difference of zero. Missing versions remain errors in the fixed denominator.

\begin{center}
\centering
\scriptsize
\setlength{\tabcolsep}{4.2pt}
\begin{tabular}{lccc}
\toprule
\textbf{Split / analysis} & \textbf{ReWeight} & \textbf{Calibrated Struct} & \textbf{Original Struct} \\
\midrule
Train & \CEightTrain & \CEightTrainCalibrated & 19/32 (59.4\%) \\
Train LOSO & \CEightLOSO & -- & -- \\
Dev confirmation & \CEightDev & \CEightDevCalibrated & 7/15 (46.7\%) \\
Extension & \CEightExtension & \CEightExtensionCalibrated & \CEightExtensionOriginal \\
\bottomrule
\end{tabular}
\captionof{table}{Offline ReWeight aggregation audit. Calibrated Struct uses the original Struct weights with the same forced-choice rule, whereas Original Struct retains the published tie threshold. The frozen L1/L2/L3 weights for ReWeight are \CEightWeights.}
\label{tab:c8-offline}
\end{center}

On the extension set, ReWeight improves over Calibrated Struct by \CEightGain{} percentage points. The 10,000-sample clustered confidence interval for this difference is \CEightBootstrapCI{} percentage points, and ReWeight does not perform worse on \CEightNonworse{} series. The 3-point gate therefore returns \texttt{\CEightStatus}. The release includes the complete diagnostic trace, including the weights, data splits, and byte-identical constraint records. All \CEightConstraintRecords{} constraint records are identical. The protocol authorizes no paid calls, and none are made (\CEightPaidCalls). The result indicates that retuning only the L1/L2/L3 aggregation weights yields no gain in hidden-test preference accuracy.

\section{Explanation-Reception Audits}
\label{supp:audits}

Two blinded audits evaluate the reception of judge rationales: a primary audit with an out-of-panel cohort and a replication with the original annotators. Both audits evaluate the delivered rationale bundle and are analyzed separately from pairwise decision accuracy.

\subsection{Out-of-Panel Audit (Primary)}

Five reviewers who did not participate in the original annotation panel complete a two-stage blinded review. Three report professional experience related to drama, and two report no such experience. Stage 1 records judgments of reasonableness and evidence support before Stage 2 reveals the five preserved human reasons. The audit includes 20 reviewable within-panel disagreements from the extension and all 12 within-panel disagreements from the original hidden test set, for a total of \IndependentAuditPairs{} pairs from \IndependentAuditSeries{} series. One additional extension disagreement, DGX-101, is excluded under the frozen reviewability rule because Flat has no complete, validated output for side B after the fixed API budget is exhausted. The failed call remains recorded as an error in the frozen run ledger and is not retried or replaced; like all within-panel disagreement pairs, DGX-101 lies outside the preference-accuracy denominator.

For each configuration, reasonableness and evidence support are evaluated on \IndependentAuditPairs{} configuration--pair majority units. Reason coverage is evaluated on \IndependentAuditCoverageUnits{} configuration--pair--reason majority units. A majority requires at least three of five ``yes'' votes. Votes of ``no'' and ``uncertain'' are counted against the majority criterion. The five reviewer records for each unit are treated as correlated observations.

\begin{center}
\centering
\scriptsize
\setlength{\tabcolsep}{3.2pt}
\begin{tabular}{lccc}
\toprule
\textbf{Config.} & \textbf{Reasonable} & \textbf{Evidence} & \textbf{Reason coverage} \\
\midrule
ZS & \auditcell{\IndCOneReason}{\IndCOneReasonWilson} & \auditcell{\IndCOneEvidence}{\IndCOneEvidenceWilson} & \auditcell{\IndCOneCoverage}{\IndCOneCoverageWilson} \\
Flat & \auditcell{\IndCThreeReason}{\IndCThreeReasonWilson} & \auditcell{\IndCThreeEvidence}{\IndCThreeEvidenceWilson} & \auditcell{\IndCThreeCoverage}{\IndCThreeCoverageWilson} \\
Struct & \auditcell{\IndCFourReason}{\IndCFourReasonWilson} & \auditcell{\IndCFourEvidence}{\IndCFourEvidenceWilson} & \auditcell{\IndCFourCoverage}{\IndCFourCoverageWilson} \\
\bottomrule
\end{tabular}
\captionof{table}{Results of the blinded out-of-panel audit (percent). The first line of each cell reports the majority estimate with a series-clustered 95\% confidence interval. The second line reports a Wilson interval.}
\label{tab:independent-rationale-audit}
\end{center}

For ZS, Flat, and Struct, the majority analysis yields \IndCOneReasonCount{}, \IndCThreeReasonCount{}, and \IndCFourReasonCount{} reasonableness decisions, respectively; \IndCOneEvidenceCount{}, \IndCThreeEvidenceCount{}, and \IndCFourEvidenceCount{} evidence-support decisions; and \IndCOneCoverageCount{}, \IndCThreeCoverageCount{}, and \IndCFourCoverageCount{} reference-reason matches.

\begin{center}
\centering
\scriptsize
\setlength{\tabcolsep}{3.2pt}
\begin{tabular}{lrrrr}
\toprule
\textbf{Config.} & \textbf{Sides} & \textbf{Completion tokens} & \textbf{Evidence spans} & \textbf{Evidence chars} \\
\midrule
ZS & \RichCOneSuccess & \RichCOneCompletionMedian{} [\RichCOneCompletionIQR] & \RichCOneSpansMedian{} [\RichCOneSpansIQR] & \RichCOneCharsMedian{} [\RichCOneCharsIQR] \\
Flat & \RichCThreeSuccess & \RichCThreeCompletionMedian{} [\RichCThreeCompletionIQR] & \RichCThreeSpansMedian{} [\RichCThreeSpansIQR] & \RichCThreeCharsMedian{} [\RichCThreeCharsIQR] \\
Struct & \RichCFourSuccess & \RichCFourCompletionMedian{} [\RichCFourCompletionIQR] & \RichCFourSpansMedian{} [\RichCFourSpansIQR] & \RichCFourCharsMedian{} [\RichCFourCharsIQR] \\
\bottomrule
\end{tabular}
\captionof{table}{Output richness for the 64 validated side-level outputs used to construct the 32 audit cards for each configuration. Entries report medians with inclusive interquartile ranges. The audit compares complete output bundles that differ in length and evidence density.}
\label{tab:audit-output-richness}
\end{center}

\begin{figure}[t]
  \centering
  \includegraphics[width=\linewidth]{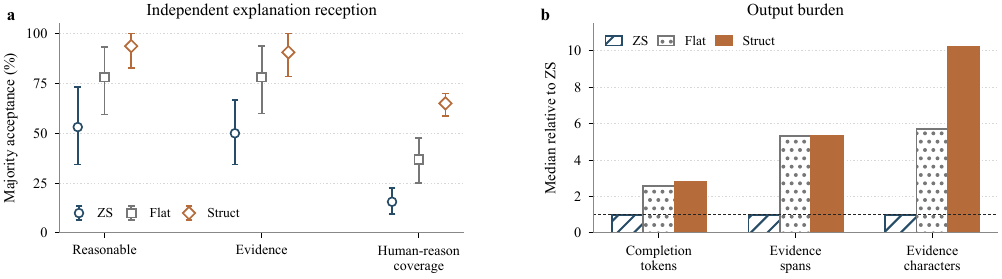}
  \caption{Rationale-reception analysis. Error bars show series-clustered 95\% confidence intervals, and output-burden bars show medians relative to ZS. Struct receives higher acceptance rates and produces longer outputs with more evidence than the other evaluated configurations. The reception results therefore describe the complete output bundle, including its structure, localization requirements, and length.}
  \label{fig:s1-explanation-reception}
\end{figure}

The following descriptive rates preserve the correlated reviewer records and stratify them by reported professional background.

\begin{center}
\centering
\scriptsize
\setlength{\tabcolsep}{3.0pt}
\begin{tabular}{llrrr}
\toprule
\textbf{Config.} & \textbf{Reviewer group} & \textbf{Reason.} & \textbf{Evidence} & \textbf{Coverage} \\
\midrule
ZS & Professional & \IndCOneProfReasonRate & \IndCOneProfEvidenceRate & \IndCOneProfCoverageRate \\
   & Nonprofessional & \IndCOneNonProfReasonRate & \IndCOneNonProfEvidenceRate & \IndCOneNonProfCoverageRate \\
   & All & \IndCOneAllReasonRate & \IndCOneAllEvidenceRate & \IndCOneAllCoverageRate \\
Flat & Professional & \IndCThreeProfReasonRate & \IndCThreeProfEvidenceRate & \IndCThreeProfCoverageRate \\
   & Nonprofessional & \IndCThreeNonProfReasonRate & \IndCThreeNonProfEvidenceRate & \IndCThreeNonProfCoverageRate \\
   & All & \IndCThreeAllReasonRate & \IndCThreeAllEvidenceRate & \IndCThreeAllCoverageRate \\
Struct & Professional & \IndCFourProfReasonRate & \IndCFourProfEvidenceRate & \IndCFourProfCoverageRate \\
   & Nonprofessional & \IndCFourNonProfReasonRate & \IndCFourNonProfEvidenceRate & \IndCFourNonProfCoverageRate \\
   & All & \IndCFourAllReasonRate & \IndCFourAllEvidenceRate & \IndCFourAllCoverageRate \\
\bottomrule
\end{tabular}
\captionof{table}{Descriptive rates of ``yes'' responses (percent) across correlated reviewer records, stratified by reported professional background.}
\label{tab:independent-groups}
\end{center}

\subsection{Within-Panel Replication}

The same five original annotators previously completed a two-stage blinded review of six anonymous system outputs for each of the 12 original disagreement pairs, using the same record-then-reveal protocol. This design produces 72 configuration--pair majority units for each Stage 1 outcome and 360 configuration--pair--reason majority units for coverage. Repeated records from the same reviewer are treated as correlated observations. The out-of-panel cohort described above provides the primary rationale-reception audit.

\begin{center}
\centering
\scriptsize
\setlength{\tabcolsep}{3.2pt}
\begin{tabular}{lccc}
\toprule
\textbf{Config.} & \textbf{Reasonable} & \textbf{Evidence} & \textbf{Reason coverage} \\
\midrule
ZS & \auditcell{\COneReason}{\COneReasonWilson} & \auditcell{\COneEvidence}{\COneEvidenceWilson} & \auditcell{\COnePlural}{\COnePluralWilson} \\
Ext & \auditcell{\CTwoReason}{\CTwoReasonWilson} & \auditcell{\CTwoEvidence}{\CTwoEvidenceWilson} & \auditcell{\CTwoPlural}{\CTwoPluralWilson} \\
Flat & \auditcell{\CThreeReason}{\CThreeReasonWilson} & \auditcell{\CThreeEvidence}{\CThreeEvidenceWilson} & \auditcell{\CThreePlural}{\CThreePluralWilson} \\
Struct & \auditcell{\CFourReason}{\CFourReasonWilson} & \auditcell{\CFourEvidence}{\CFourEvidenceWilson} & \auditcell{\CFourPlural}{\CFourPluralWilson} \\
Flat+HL & \auditcell{\CSixReason}{\CSixReasonWilson} & \auditcell{\CSixEvidence}{\CSixEvidenceWilson} & \auditcell{\CSixPlural}{\CSixPluralWilson} \\
Struct+HL-c & \auditcell{\CSevenCReason}{\CSevenCReasonWilson} & \auditcell{\CSevenCEvidence}{\CSevenCEvidenceWilson} & \auditcell{\CSevenCPlural}{\CSevenCPluralWilson} \\
\bottomrule
\end{tabular}
\captionof{table}{Results of the blinded within-panel audit (percent). The first line of each cell reports the point estimate with a series-clustered 95\% confidence interval, and the second line reports the binomial Wilson interval. Reasonableness and evidence support use 12 configuration--pair units from nine series. Coverage uses 60 configuration--pair--reason units from the same nine series.}
\label{tab:human-rationale-audit}
\end{center}

Struct receives a majority reasonableness decision for 12 of 12 pairs and a majority evidence-support decision for 10 of 12 pairs. Struct+HL-c covers 22 of the 60 preserved reasons. These results describe the reception of rationales by the original panel and follow the same overall direction as the out-of-panel audit in Table~\ref{tab:independent-rationale-audit}.

\section{Corpus and Model Audit}
\label{supp:corpus}

Anonymous labels identify the four heterogeneous generation sources. Every source uses the same generation prompt, temperature of 0.8, top-$p$ of 0.95, and 5K output limit, with reasoning disabled.

\begin{center}
\scriptsize
\setlength{\tabcolsep}{2.3pt}
\begin{tabular}{clrr}
\toprule
\textbf{Source} & \textbf{Model} & \textbf{Median} & \textbf{Mean} \\
\midrule
A & \texttt{qwen3.7-max-2026-06-08} & 43 & 43.2 \\
B & \texttt{deepseek-v4-pro} & 57 & 55.3 \\
C & \texttt{glm-4.7} & 36 & 37.1 \\
D & \texttt{qwen3.6-flash-2026-04-16} & 54 & 53.2 \\
\bottomrule
\end{tabular}
\captionof{table}{Audited nonblank lines per episode.}
\label{tab:source-length}
\end{center}

The frozen pool contains 200 episodes from each source and no duplicate groups. Correlation between score and length is used to quantify length sensitivity independently of source identity.

The rubric optimizer is \texttt{deepseek-v4-pro} and is accessed through the official DeepSeek API. The primary judge is \texttt{glm-5.2}, and the secondary judge is the dated \texttt{qwen3.7-plus-2026-05-26} snapshot\ifKimiCrossJudgeResults. A separate frozen addendum uses \texttt{kimi-k2.6} with reasoning enabled\fi. The judges use the mainland China endpoint of Alibaba Cloud Bailian with a 24K output limit. The logs record the requested and returned model identifiers, endpoint, reasoning mode, token counts, latency, finish reason, retries, and call time. GLM is used for rubric-error analysis, rubric updates, and development-set selection. The frozen rubrics are subsequently evaluated on Qwen\ifKimiCrossJudgeResults\ and Kimi\fi\ without adaptation. This design evaluates one-way transfer of rubrics formed or selected with GLM.

\section{Historical 25-Pair Pilots}
\label{supp:pilots}

The two studies in this section precede the unified comparison on 114 preference pairs and 35 constraint interventions. They are retained as frozen pilot diagnostics in separately named artifact families.

\subsection{One-Way Cross-Judge Transfer}

Table~\ref{tab:cross-judge-absolute} reports the common-slice diagnostic for each unique internal configuration under every completed judge. Each panel-supermajority estimate uses 25 pairs from 15 series, and each constraint estimate uses 35 interventions from 16 series. Changing the judge alters both the absolute estimates and the system ordering. For example, the accuracy of Struct changes from \CrossGLMCFourConsensus\% with GLM to \CrossQwenCFourConsensus\% with Qwen, whereas the accuracy of Struct+HL-c changes from \CrossGLMCSevenCConsensus\% to \CrossQwenCSevenCConsensus\%. These differences quantify judge dependence. The primary ordering on 114 preference pairs is based on the unified full-denominator baselines in Section~\ref{supp:baselines}.

\begin{table}[t]
\centering
\scriptsize
\captionsetup{skip=4pt}
\renewcommand{\arraystretch}{0.80}
\setlength{\tabcolsep}{4.0pt}
\begin{tabular}{lrrrr}
\toprule
\multicolumn{5}{c}{\textbf{Primary GLM judge}} \\
\cmidrule(lr){1-5}
\textbf{Config.} & \textbf{Cons.} & \textbf{F1} & \textbf{D+L} & \textbf{FPR} \\
\midrule
ZS & \metriccell{\CrossGLMCOneConsensus}{\CrossGLMCOneConsensusCI} & \metriccell{\CrossGLMCOneConstraint}{\CrossGLMCOneConstraintCI} & \metriccell{\CrossGLMCOneLocation}{\CrossGLMCOneLocationCI} & \metriccell{\CrossGLMCOneFPR}{\CrossGLMCOneFPRCI} \\
Ext & \metriccell{\CrossGLMCTwoConsensus}{\CrossGLMCTwoConsensusCI} & \metriccell{\CrossGLMCTwoConstraint}{\CrossGLMCTwoConstraintCI} & \metriccell{\CrossGLMCTwoLocation}{\CrossGLMCTwoLocationCI} & \metriccell{\CrossGLMCTwoFPR}{\CrossGLMCTwoFPRCI} \\
Flat & \metriccell{\CrossGLMCThreeConsensus}{\CrossGLMCThreeConsensusCI} & \metriccell{\CrossGLMCThreeConstraint}{\CrossGLMCThreeConstraintCI} & \metriccell{\CrossGLMCThreeLocation}{\CrossGLMCThreeLocationCI} & \metriccell{\CrossGLMCThreeFPR}{\CrossGLMCThreeFPRCI} \\
Struct & \metriccell{\CrossGLMCFourConsensus}{\CrossGLMCFourConsensusCI} & \metriccell{\CrossGLMCFourConstraint}{\CrossGLMCFourConstraintCI} & \metriccell{\CrossGLMCFourLocation}{\CrossGLMCFourLocationCI} & \metriccell{\CrossGLMCFourFPR}{\CrossGLMCFourFPRCI} \\
Flat+HL & \metriccell{\CrossGLMCSixConsensus}{\CrossGLMCSixConsensusCI} & \metriccell{\CrossGLMCSixConstraint}{\CrossGLMCSixConstraintCI} & \metriccell{\CrossGLMCSixLocation}{\CrossGLMCSixLocationCI} & \metriccell{\CrossGLMCSixFPR}{\CrossGLMCSixFPRCI} \\
Struct+HL-c & \metriccell{\CrossGLMCSevenCConsensus}{\CrossGLMCSevenCConsensusCI} & \metriccell{\CrossGLMCSevenCConstraint}{\CrossGLMCSevenCConstraintCI} & \metriccell{\CrossGLMCSevenCLocation}{\CrossGLMCSevenCLocationCI} & \metriccell{\CrossGLMCSevenCFPR}{\CrossGLMCSevenCFPRCI} \\
\midrule
\multicolumn{5}{c}{\textbf{Zero-adaptation Qwen judge}} \\
\cmidrule(lr){1-5}
\textbf{Config.} & \textbf{Cons.} & \textbf{F1} & \textbf{D+L} & \textbf{FPR} \\
\midrule
ZS & \metriccell{\CrossQwenCOneConsensus}{\CrossQwenCOneConsensusCI} & \metriccell{\CrossQwenCOneConstraint}{\CrossQwenCOneConstraintCI} & \metriccell{\CrossQwenCOneLocation}{\CrossQwenCOneLocationCI} & \metriccell{\CrossQwenCOneFPR}{\CrossQwenCOneFPRCI} \\
Ext & \metriccell{\CrossQwenCTwoConsensus}{\CrossQwenCTwoConsensusCI} & \metriccell{\CrossQwenCTwoConstraint}{\CrossQwenCTwoConstraintCI} & \metriccell{\CrossQwenCTwoLocation}{\CrossQwenCTwoLocationCI} & \metriccell{\CrossQwenCTwoFPR}{\CrossQwenCTwoFPRCI} \\
Flat & \metriccell{\CrossQwenCThreeConsensus}{\CrossQwenCThreeConsensusCI} & \metriccell{\CrossQwenCThreeConstraint}{\CrossQwenCThreeConstraintCI} & \metriccell{\CrossQwenCThreeLocation}{\CrossQwenCThreeLocationCI} & \metriccell{\CrossQwenCThreeFPR}{\CrossQwenCThreeFPRCI} \\
Struct & \metriccell{\CrossQwenCFourConsensus}{\CrossQwenCFourConsensusCI} & \metriccell{\CrossQwenCFourConstraint}{\CrossQwenCFourConstraintCI} & \metriccell{\CrossQwenCFourLocation}{\CrossQwenCFourLocationCI} & \metriccell{\CrossQwenCFourFPR}{\CrossQwenCFourFPRCI} \\
Flat+HL & \metriccell{\CrossQwenCSixConsensus}{\CrossQwenCSixConsensusCI} & \metriccell{\CrossQwenCSixConstraint}{\CrossQwenCSixConstraintCI} & \metriccell{\CrossQwenCSixLocation}{\CrossQwenCSixLocationCI} & \metriccell{\CrossQwenCSixFPR}{\CrossQwenCSixFPRCI} \\
Struct+HL-c & \metriccell{\CrossQwenCSevenCConsensus}{\CrossQwenCSevenCConsensusCI} & \metriccell{\CrossQwenCSevenCConstraint}{\CrossQwenCSevenCConstraintCI} & \metriccell{\CrossQwenCSevenCLocation}{\CrossQwenCSevenCLocationCI} & \metriccell{\CrossQwenCSevenCFPR}{\CrossQwenCSevenCFPRCI} \\
\ifKimiCrossJudgeResults
\midrule
\multicolumn{5}{c}{\textbf{Zero-adaptation Kimi judge}} \\
\cmidrule(lr){1-5}
\textbf{Config.} & \textbf{Cons.} & \textbf{F1} & \textbf{D+L} & \textbf{FPR} \\
\midrule
ZS & \metriccell{\CrossKimiCOneConsensus}{\CrossKimiCOneConsensusCI} & \metriccell{\CrossKimiCOneConstraint}{\CrossKimiCOneConstraintCI} & \metriccell{\CrossKimiCOneLocation}{\CrossKimiCOneLocationCI} & \metriccell{\CrossKimiCOneFPR}{\CrossKimiCOneFPRCI} \\
Ext & \metriccell{\CrossKimiCTwoConsensus}{\CrossKimiCTwoConsensusCI} & \metriccell{\CrossKimiCTwoConstraint}{\CrossKimiCTwoConstraintCI} & \metriccell{\CrossKimiCTwoLocation}{\CrossKimiCTwoLocationCI} & \metriccell{\CrossKimiCTwoFPR}{\CrossKimiCTwoFPRCI} \\
Flat & \metriccell{\CrossKimiCThreeConsensus}{\CrossKimiCThreeConsensusCI} & \metriccell{\CrossKimiCThreeConstraint}{\CrossKimiCThreeConstraintCI} & \metriccell{\CrossKimiCThreeLocation}{\CrossKimiCThreeLocationCI} & \metriccell{\CrossKimiCThreeFPR}{\CrossKimiCThreeFPRCI} \\
Struct & \metriccell{\CrossKimiCFourConsensus}{\CrossKimiCFourConsensusCI} & \metriccell{\CrossKimiCFourConstraint}{\CrossKimiCFourConstraintCI} & \metriccell{\CrossKimiCFourLocation}{\CrossKimiCFourLocationCI} & \metriccell{\CrossKimiCFourFPR}{\CrossKimiCFourFPRCI} \\
Flat+HL & \metriccell{\CrossKimiCSixConsensus}{\CrossKimiCSixConsensusCI} & \metriccell{\CrossKimiCSixConstraint}{\CrossKimiCSixConstraintCI} & \metriccell{\CrossKimiCSixLocation}{\CrossKimiCSixLocationCI} & \metriccell{\CrossKimiCSixFPR}{\CrossKimiCSixFPRCI} \\
Struct+HL-c & \metriccell{\CrossKimiCSevenCConsensus}{\CrossKimiCSevenCConsensusCI} & \metriccell{\CrossKimiCSevenCConstraint}{\CrossKimiCSevenCConstraintCI} & \metriccell{\CrossKimiCSevenCLocation}{\CrossKimiCSevenCLocationCI} & \metriccell{\CrossKimiCSevenCFPR}{\CrossKimiCSevenCFPRCI} \\
\fi
\bottomrule
\end{tabular}
\caption{Absolute estimates for the 25-pair one-way transfer pilot (percent), with series-clustered 95\% confidence intervals in brackets. ``Cons.'' denotes agreement with the within-panel supermajority reference, D+L denotes joint detection--localization, and FPR denotes the false-positive rate.}
\label{tab:cross-judge-absolute}
\end{table}

The wide confidence intervals and low rank correlations, $-0.19$, $0.26$, and $0.39$ for panel-supermajority accuracy, F1, and D+L, respectively, indicate substantial dependence of the single-call ordering on the judge pipeline.

\subsection{External Baseline Pilot}

Before any calls are made on the common pilot subset, the signed freeze fixes the task adapters, prompts, calibration procedure, runner, and analysis code. ToW-adapted follows the published 5:3:1 aggregation of content, format, and impression scores. It returns evidence for its preference decision but does not produce fatal-gate or episode-localization outputs. RULERS-adapted uses a fixed drama checklist, extractive-evidence validation, and a decision boundary selected on 15 development pairs. It completes all 35 constraint pairs and 24 of 25 panel-supermajority pairs. DG-085 exhausts the frozen retry policy and remains in the denominator as an incorrect abstention. Results from this 25-pair pilot are excluded from the ranking on the full set of 114 preference pairs.

\section{Rubric Representation and HL Controls}
\label{supp:rubric}

The detailed seed contains five fatal gates and 16 quality dimensions. L1 covers dialogue naturalness, character voice, subtext, exposition efficiency, and agreement between dialogue and emotion. L2 covers opening hooks, escalation, setup--payoff completeness, scene functionality, episode-end hooks, and pacing density. L3 covers cross-episode continuity, character scheduling, relationship progression, knowledge boundaries, and focus on the main plot.

Each criterion contains a stable identifier, a natural-language description, scope, target unit, required evidence locations, specification anchors, cross-unit dependencies, and score boundaries. The matched flat baseline expresses the same leaf-level content and evidence requirements in prose. The structured version adds machine-readable fields and requires localized evidence.

\paragraph{Heuristic rubric learning (HL) controls.}
The learned configurations are fine-grained structured learning (Struct+HL-f), coarse-grained structured learning (Struct+HL-c), and flat learning (Flat+HL). All three use \texttt{max-iters=10}, \texttt{patience=2}, six workers, and a two-series support rule. Training labels support rubric updates, development labels determine rubric selection, and test tasks remain sealed. Within-panel disagreements are used only in the post-freeze rationale analysis. Before execution, the signed manifest records the rubrics, prompts, configuration, code, exclusions, and hashes.

\end{document}